\documentclass[lettersize,journal,twoside]{IEEEtran}
\usepackage{amsmath,amsfonts}
\usepackage{algorithm}
\usepackage{algpseudocode}
\usepackage{array}
\usepackage[caption=false,font=footnotesize]{subfig}
\usepackage{textcomp}
\usepackage{stfloats}
\usepackage{url}
\usepackage{verbatim}
\usepackage{graphicx}
\usepackage{cite}
\usepackage{lipsum}
\usepackage{color}
\usepackage[hidelinks, colorlinks=true, linkcolor=blue, citecolor=blue, urlcolor=blue]{hyperref}
\newif\ifarxiv
\arxivtrue 

\ifarxiv
\usepackage{tikz}
\fi

\begin{document}
\bstctlcite{IEEEexample:BSTcontrol}
\newcommand{\eg}[1]{\textit{e.g.,}~{#1}}

\title{Learning Autonomous Surgical Irrigation and Suction with the da Vinci Research Kit Using Reinforcement Learning}

\author{Yafei Ou$^{1}$,~\IEEEmembership{Graduate Student Member,~IEEE,}
        Mahdi Tavakoli$^{1,2}$,~\IEEEmembership{Senior Member,~IEEE}
\thanks{This research was supported by the Canada Foundation for Innovation (CFI), the Natural Sciences and Engineering Research Council (NSERC) of Canada, the Canadian Institutes of Health Research (CIHR), Alberta Innovates, the China Scholarship Council (CSC), and Alberta Advanced Education. \textit{(Corresponding author: Yafei Ou.)}}
\thanks{$^{1}$Yafei Ou and Mahdi Tavakoli are with the Department of Electrical and Computer Engineering, University of Alberta, Edmonton, Alberta, Canada (e-mail: {\tt\small \{yafei.ou, mahdi.tavakoli\}@ualberta.ca}).}
\thanks{$^{2}$Mahdi Tavakoli is also with the Department of Biomedical Engineering, University of Alberta, Edmonton, Alberta, Canada.}%
}



\maketitle

\ifarxiv
\begin{tikzpicture}[remember picture, overlay]
\node [align=left, xshift=10cm, yshift=-0.8cm] at (current page.north west) 
{
\begin{minipage}{19cm} 
\footnotesize
\textcopyright 2025 IEEE.  Personal use of this material is permitted.  Permission from IEEE must be obtained for all other uses, in any current or future media, including reprinting/republishing this material for advertising or promotional purposes, creating new collective works, for resale or redistribution to servers or lists, or reuse of any copyrighted component of this work in other works.
\end{minipage}
};
\end{tikzpicture}
\fi

\begin{abstract}
The irrigation-suction process is a common procedure to rinse and clean up the surgical field in minimally invasive surgery (MIS).
In this process, surgeons first irrigate liquid, typically saline, into the surgical scene for rinsing and diluting the contaminant, and then suction the liquid out of the surgical field.
While recent advances have shown promising results in the application of reinforcement learning (RL) for automating surgical subtasks, fewer studies have explored the automation of fluid-related tasks.
In this work, we explore the automation of both steps in the irrigation-suction procedure and train two vision-based RL agents to complete irrigation and suction autonomously.
To achieve this, a platform is developed for creating simulated surgical robot learning environments and for training agents, and two simulated learning environments are built for irrigation and suction with visually plausible fluid rendering capabilities.
With techniques such as domain randomization (DR) and imitation learning, two agents are trained in the simulator and transferred to the real world.
Individual evaluations of both agents show satisfactory real-world results.
With an initial amount of around 5 grams of contaminants, the irrigation agent ultimately achieved an average of 2.21 grams remaining after a manual suction.
As a comparison, fully manual operation by a human results in 1.90 grams remaining.
The suction agent achieved 2.64 and 2.24 grams of liquid remaining across two trial groups with more than 20 and 30 grams of initial liquid in the container.
Fully autonomous irrigation-suction trials reduce the contaminant in the container from around 5 grams to an average of 2.42 grams, although yielding a higher total weight remaining (4.40) due to residual liquid not suctioned.
Further information about the project is available at \url{https://tbs-ualberta.github.io/CRESSim/}.

\textit{Note to Practitioners}---The irrigation-suction process is a surgical procedure for rinsing and cleaning surgical fields. This work tackles automating the process to reduce the workload for surgeons. Our approach is based on two customized simulation environments that can simulate the irrigation and suction process realistically. Two autonomous agents are trained using robot learning approaches in the environments for completing irrigation and suction, respectively, and then transferred to the real world.
The agents autonomously control the surgical robot by interpreting the images captured from an RGB camera and the robot's current state, and generate joint movements of the robot.
This approach has been tested in physical settings, showing promising results in terms of the individual and combined performance of the agents in executing the full irrigation-suction process.
Future work is needed to extend the approach to more practical surgical settings, evaluate the performance under diverse conditions, and enhance integration with existing surgical robot platforms.
\end{abstract}

\begin{IEEEkeywords}
Medical robots and systems, surgical robotics, fluid simulation, reinforcement learning.
\end{IEEEkeywords}

\section{Introduction}
\IEEEPARstart{A} growing interest has been shown in recent years in achieving autonomous or semi-autonomous subtask execution in surgeries using surgical robotic systems.
These include tissue cutting \cite{nguyen2019manipulating,nguyen2019new,shahkoo2023deep}, tissue manipulation \cite{tagliabue2020soft,scheikl2022sim,shahkoo2023autonomous,ou2023sim}, suturing \cite{varier2020collaborative,chiu2021bimanual,bendikas2023learning}, and others, with the overall goal of reducing the workload and enhancing surgeons' skills when performing specific subtasks, achieving what is termed \textit{augmented dexterity}---the enhancement of a surgeon’s capabilities through robotic assistance under supervision \cite{goldberg2024augmented}.
Increasing numbers of these studies are utilizing machine learning approaches, particularly robot learning techniques such as reinforcement learning (RL) and imitation learning (IL), due to their generalizability and reduced manual effort when compared with traditional motion planning algorithms, thanks to their data-driven nature.

While these studies show promising results, many rely on manually extracted feature vectors from images as input to the policy, such as tooltip positions and key point locations extracted from markers placed on the surgical instrument or tissue. Although technically feasible, this can be less robust in realistic surgical scenarios, such as when the lighting conditions change and when smoke is present due to energy-based operations.
Therefore, relying directly on the raw visual observation captured from the endoscopic camera as input to the policy is usually more desirable, due to the elimination of manual feature extraction and the potential of obtaining an end-to-end policy that takes the sensory input (\eg{image and sensor reading}) and directly produces the desired actions.

Training vision-based agents that take image observations has been extensively explored in general robot manipulation and navigation. A number of these studies rely on a simulation-to-reality (sim-to-real, \textit{sim2real}) approach, where the agent is first trained in a simulated environment that synthesizes the image observation, and then transferred to the real world.
Several recent studies in surgical robotics have focused on this aspect as well. In~\cite{gondokaryono2023learning}, an agent is trained to roll a block using the surgical robot in a simulator that synthesizes $64\times64$ RGB image observations with domain randomization (DR), where the environment parameters such as lighting and camera positions are randomized, and then transferred to the real world directly. Using a similar approach, Haiderbhai \textit{et al.}~obtained a vision-based policy for rope cutting \cite{haiderbhai2024sim2real}. In~\cite{scheikl2022sim}, the authors achieved sim-to-real transfer for autonomous tissue manipulation by utilizing domain adaptation (DA) using a contrastive generative adversarial network (GAN).

However, one major challenge when applying sim-to-real approaches to surgical task automation is the need to simulate realistic surgical scenes and operations.
Unlike simulators for everyday and industrial robotic tasks, which typically involve mostly rigid objects governed by basic principles of mechanics, surgical simulations require a higher level of complexity to accurately model soft tissues, fluids, and dynamic interactions such as cutting and cauterization.
The simulation of such physical models and behaviors is already a considerable challenge involving computational solid and fluid mechanics, but rendering visually realistic images for fluid in the simulator to support the training of vision-based agents adds an additional layer of complexity.

These complexities are particularly evident in the underexplored area of surgical irrigation and suction.
During surgery, blood, debris, and other contaminants are frequently present as a result of procedures like cutting, cauterization, and tissue manipulation \cite{Milsom2006}.
To maintain a clear surgical field, surgeons may follow a two-step process: first, they irrigate the area with a sterile solution, typically saline, to rinse or dilute the contaminants; then, suction is used to remove the fluid along with any remaining debris.
This process is usually carried out using an irrigation-suction instrument in robotic surgery, which allows the surgeon to both introduce and remove fluids.
However, this usually involves a considerable amount of time and effort \cite{tsao2023saline}, and automating this process may help reduce the workload of surgeons.

Developing simulated learning environments for the task is challenging, as it involves fluid dynamics, which is more computationally complex than rigid body dynamics, and the corresponding interactions for irrigating and suctioning fluids.
Fluid mixing should also be considered, as the irrigated solution will be mixed with the existing contaminants.
Additionally, developing a vision-based policy involves a greater degree of complexity since the rendering of fluids while mixing them is also required to synthesize visually plausible images.

Existing studies that attempt to automate this task primarily focus on the suction phase, particularly blood suction \cite{lai2021verticalized,richter2021autonomous,huang2021model,ou2024autonomous,zargarzadeh2025decision}.
In \cite{richter2021autonomous}, the authors employ optical flow to track blood flow in real-time and generate a suction trajectory using a computational motion planner to efficiently remove blood while moving upstream toward the bleeding point.
Huang \textit{et~al.\ }implemented a differentiable position-based fluids (PBF) model of the blood and used model predictive control (MPC) to generate a suction trajectory.
In \cite{ou2024autonomous}, a simulated blood suction environment is built based on PBF and is used to train an RL agent that takes a binary image mask of the blood region as input, and outputs the 2D motion of the suction tool. However, manual feature extraction is still needed to obtain a binary mask representing the area of the blood.
To the best of the authors' knowledge, none of the existing studies have developed an agent relying on raw RGB images without manual feature extraction, and the automation of irrigation has not been explored in the literature.

\begin{figure*}[ht]
\centering
    \includegraphics[width=0.9\textwidth]{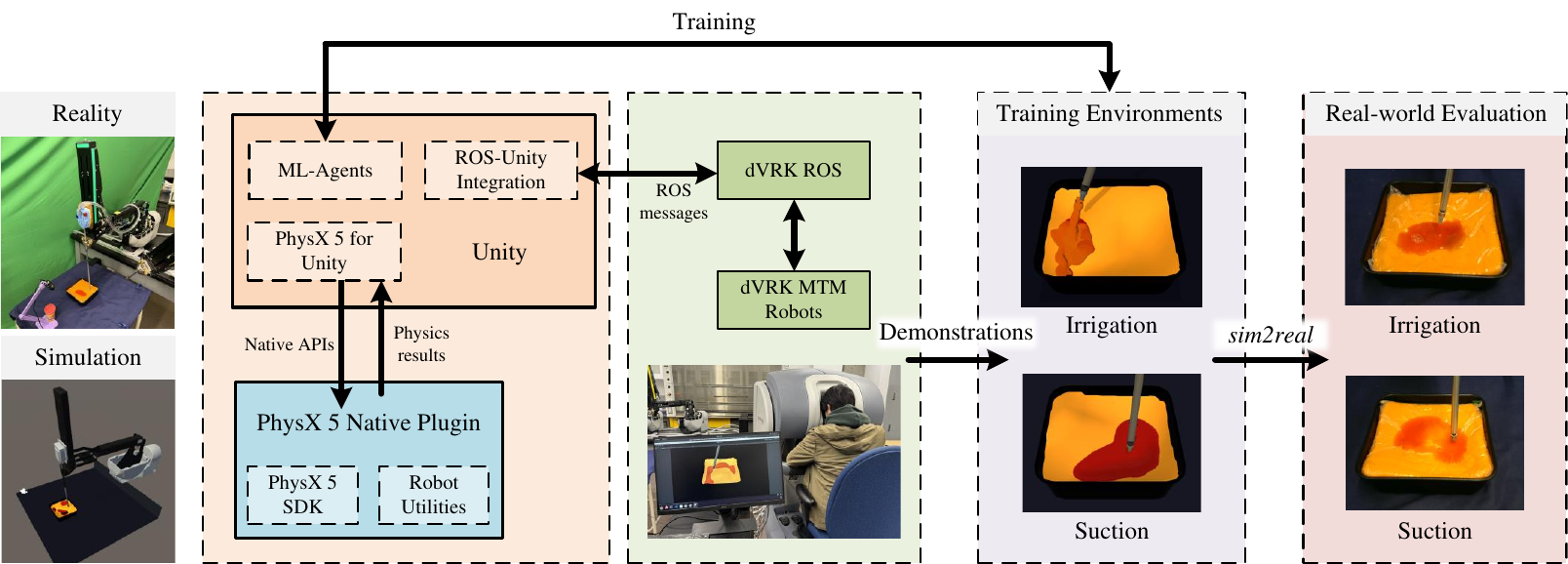}%
\caption{Overall framework.}
\label{fig:framework}
\end{figure*}

In this work, we develop vision-based agents that autonomously complete both steps during surgical irrigation and suction. To achieve this, we build a novel surgical robot learning platform for the da Vinci Research Kit (dVRK) \cite{kazanzides2014open} that integrates robot learning capabilities and visually realistic fluid simulations and rendering. We design irrigation and suction learning environments in the simulator that resemble the real-world setup. Vision-based agents are trained in the simulator with domain-randomized parameters and carefully designed rewards, learning curricula, and human demonstrations, and are then transferred to the real world.
This approach is similar to the one used in \cite{haiderbhai2024sim2real}, where a vision-based rope-cutting model is obtained in a sim-to-real manner.
The main contributions of this work are as follows:
\begin{itemize}
    \item We propose CRESSim-ML, a surgical robot learning platform for the dVRK.
    Built on our previous work (CRESSim) \cite{ou2024realistic}, the platform additionally incorporates parallel training environment capabilities, and allows collecting human demonstrations through teleoperation using the actual dVRK Master Tool Manipulator (MTM).
    \item Based on CRESSim-ML and Unity's ML-Agents, simulated learning environments are created for irrigation and suction using the Patient Side Manipulator (PSM) from the dVRK and the EndoWrist Suction/Irrigator. Screen-space fluid rendering is implemented and adapted to simulate realistic visual effects, including fluid mixing.
    \item With hand-crafted reward functions, DR, curriculum learning (CL), and IL, two vision-based agents for irrigation and suction are trained in the simulator for controlling the robot in joint space to complete the tasks.
    \item We conduct real-world experiments to validate the sim-to-real transfer of the trained agents and analyze their performance quantitatively.
\end{itemize}
To the best of the authors' knowledge, this is the first study to investigate the automation of the two-step irrigation and suction processes.
It is also one of the few studies that use a learning-based approach to automate the motion of the surgical robot in joint space based on RGB image observations without additional feature extraction.
The overall framework is shown in Fig.~\ref{fig:framework}.

\section{Related Work}
\label{sec:related_work}
\subsection{Surgical Subtask Automation and Augmented Dexterity}
A large number of recent studies aim to achieve subtask automation in surgeries using surgical robotic systems, with a long-term goal of increasing the level of autonomy in surgeries, similar to autonomous vehicles. The concept where surgical subtasks are controlled by a robot under surgeons' supervision is recently referred to as \textit{augmented dexterity} \cite{goldberg2024augmented}. Examples of these studies include tissue cutting \cite{nguyen2019manipulating,nguyen2019new,shahkoo2023deep}, tissue manipulation \cite{scheikl2022sim,shahkoo2023autonomous,ou2023sim}, suturing \cite{varier2020collaborative,chiu2021bimanual,bendikas2023learning}, blood suction \cite{lai2021verticalized,richter2021autonomous,huang2021model,ou2024autonomous,zargarzadeh2025decision}, and vessel manipulation \cite{dharmarajan2023automating}.
Numerous studies have also focused on the navigation and control of endoscopes \cite{li20223d,li2022learning,gao2022savanet}.
Many of these studies leverage RL approaches \cite{nguyen2019manipulating,nguyen2019new,shahkoo2023deep,scheikl2022sim,shahkoo2023autonomous,ou2023sim,chiu2021bimanual,bendikas2023learning,varier2020collaborative,ou2024autonomous}.
However, applying RL to surgical tasks involving fluid manipulation, such as irrigation and suction, remains relatively underexplored due to the challenges and limited software tools for simulating fluid dynamics.

\subsection{Surgical Simulation and Sim-to-Real Transfer}
Surgical task simulation is challenging compared with daily tasks, due to the multiple types of objects and manipulations involved, such as soft bodies (\eg{soft tissue}), fluids (\eg{blood and other body fluids}), burning, and cutting.
Nevertheless, its applications have been important in two major domains. The traditional usage is for training surgeons, where trainees operate in a simulated surgical environment to improve their skills. Examples of such systems include
da Vinci SimNow\footnote{\url{www.intuitive.com/en-us/products-and-services/da-vinci/learning/simnow}},
VirtaMed LaproS\footnote{\url{www.virtamed.com/en/products-and-solutions/simulators/laparos}},
and iMSTK\footnote{\url{www.imstk.org}}.

A more emerging application of surgical simulation is for training autonomous agents using robot learning methods, where agents are trained in the simulator and transferred to the real world.
Studies of this type include \cite{tagliabue2020soft,ou2023sim,chiu2021bimanual,bendikas2023learning,haiderbhai2024sim2real,ou2024autonomous}.
While it is ideal to use commercial simulators such as the da Vinci SimNow for building training environments for robot learning, it is challenging in practice due to the proprietary nature of these software systems.
Many recent studies have proposed open-source simulation environments for surgical robot learning, including dVRL \cite{richter2019open}, AMBF-RL \cite{varier2022ambf}, UnityFlexML \cite{tagliabue2020soft}, SurRoL \cite{xu2021surrol,long2023human}, LapGym \cite{scheikl2023lapgym}, Surgical Gym \cite{schmidgall2024surgical}, and ORBIT-Surgical \cite{yu2024orbit}.
Among them, a number of recent studies consider GPU-accelerated simulation, which allows large-scale parallel training and significantly boosts training efficiency and performance.
However, they are generally inferior to the commercial ones in terms of the types of objects that they can simulate. For instance, AMBF-RL is mainly used for simulating rigid bodies and some soft bodies, without the capability for simulating fluids.
Nevertheless, achieving sim-to-real transfer for irrigation-suction requires the simulation of visually plausible fluids.
There are also studies that build task-specific simulation environments, such as \cite{gondokaryono2023learning,haiderbhai2024sim2real}.

\subsection{Fluid Simulation and Manipulation}
This work is also related to recent advancements in robotic manipulation involving fluids, as well as the development of fluid simulation environments for robot learning \cite{ma2018fluid,babaians2022pournet,xian2022fluidlab}.
In \cite{ma2018fluid}, the Navier–Stokes equation is used for simulating 2D fluid flow, and an RL model is trained to learn to manipulate a rigid object using a water jet.
Babaians \textit{et al.\ }used PBF for simulating fluids \cite{babaians2022pournet}, which is used to train a robot to pour a glass of liquid.
Authors of \cite{xian2022fluidlab} considered a number of fluid manipulation tasks, such as fluid mixing and drawing latte art.
Most of these studies focus on daily tasks such as pouring and beverage mixing, which is different from surgical irrigation or suction in terms of manipulation contact.
For instance, daily tasks rarely involve fluid suction that requires the simulation of a suction force.

There are a number of real-time fluid simulation and animation techniques, such as smoothed-particle hydrodynamics (SPH), material point method (MPM), fluid implicit particle (FLIP), and PBF \cite{wang2024physics}.
Among these methods, PBF is one of the most computationally efficient and numerically stable ones due to its consideration of positional constraints instead of relying on force integration.
This makes it particularly well-suited for large-scale trial-and-error learning, enabling massive parallel training, such as in \cite{babaians2022pournet}, while avoiding pauses or time rewinds caused by numerical instability.
For fluid rendering, there are two main technical approaches: (a) mesh-based surface reconstruction, and (b) screen-space rendering.
The first approach reconstructs a surface mesh from fluid particles or level sets using techniques such as marching cubes \cite{muller2009fast}. 
However, this approach is generally more computationally expensive than the screen-space method, which approximates the fluid surface directly in image space using depth maps or normal reconstruction techniques \cite{cords2009interactive,van2009screen,xu2023anisotropic}.
This is widely used in real-time applications, often in combination with PBF to achieve visually plausible results with less computation.

\section{Simulated Learning Environments for Irrigation and Suction}
\label{sec:simulated_learning_envs}
\subsection{Fluid Simulation and Rendering}
\label{sec:fluid_sim_render}
\subsubsection{Fluid Simulation With Color Diffusion}
To simulate irrigation and suction, fluid behavior must be simulated and rendered with visual plausibility. PhysX 5 SDK provides the feature of simulating fluid particles on GPU using position-based dynamics (PBD), and more specifically, position-based fluids (PBF) \cite{macklin2013position}.
Although PhysX 5 includes an existing PBF implementation, it focuses solely on numerical computation without rendering considerations.
In this work, we further introduce a straightforward approach to fluid color mixing when using PhysX's PBF.

\begin{figure}[t]
    \centering
    \subfloat[]{%
    \includegraphics[width=0.3\columnwidth]{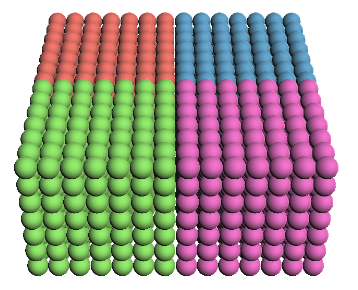}%
    }\hfil
    \subfloat[]{%
    \includegraphics[width=0.3\columnwidth]{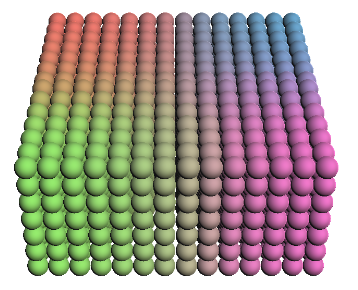}%
    }
    \caption{Particle color diffusion. (a) Initial colors; (b) Colors after 10 steps.}
    \label{fig:color_diffusion}
\end{figure}

In PBF, fluid is modeled by small particles with positions $\mathbf{x}_i$ and velocities $\mathbf{v}_i$. During each simulation step, the velocities of the particles are first predicted based on the external forces $\mathbf{f}_{ext}(\mathbf{x}_i)$, such as gravity:
\begin{equation}
\label{eqn:pbf_velocity}
    \mathbf{v}_i \gets \mathbf{v}_i + \Delta\tau \mathbf{f}_{ext}(\mathbf{x}_i),
\end{equation}
where $\Delta \tau$ is the simulation time step. The next-step positions can thus be predicted by $\mathbf{p}_i = \mathbf{x}_i + \Delta\tau \mathbf{v}_{i}$.
$\mathbf{p}_i$ is then iteratively corrected by solving positional constraints, such as collision. Finally, the velocities are updated based on the predicted position change of the particles, and the positions are set to the predicted ones:
\[
    \mathbf{v}_i \gets (\mathbf{p}_i - \mathbf{x}_i)/\Delta\tau , \quad
    \mathbf{x}_i \gets \mathbf{p}_i.
\]

As we would like to mix two types of fluids with different colors when simulating irrigation, the simulation should include color mixing and changes. It is therefore straightforward to label each particle with a color and diffuse the colors between particles based on their spatial proximity.
To achieve this, each particle is assigned to a cell in a 3D grid with uniform discrete cells based on its position, also known as spatial hashing.
For each particle, all particles in a neighboring region are evaluated, and the weighted average of the particle colors is computed to update the particle’s color. Therefore, for each particle $i$ with color $\mathbf{c}_i$,
\begin{equation}
    \mathbf{c}_i = \frac{\sum_j\mathbf{w}_{ij} * \mathbf{c}_j}{\sum_j\mathbf{w}_{ij}}.
\end{equation}
Weighting can be done based on the distance between two particles, with $\mathbf{w}_{ij} = \exp(- \left\lVert \mathbf{p}_i - \mathbf{p}_j \right\rVert / 2 \sigma^2 )$, similar to applying a Gaussian filter.
Particle velocity can be considered as an additional factor, assuming that particles with higher speed should contribute more during color diffusion:
\[
    \mathbf{w}_{ij} = \exp(- \left\lVert \mathbf{p}_i - \mathbf{p}_j \right\rVert / 2 \sigma^2 ) \cdot c \cdot \left\lVert \mathbf{v}_j \right\rVert, \; (i \neq j),
\]
where $c$ is a constant. In this way, particles that are closer to one another and have a higher velocity can influence each other's color more significantly.
This process is repeated at each step using a compute shader on the GPU. An example is shown in Fig.~\ref{fig:color_diffusion}.

\subsubsection{Fluid Rendering}
To render the particles as fluid, one common approach is screen-space fluid rendering \cite{cords2009interactive,van2009screen}, where the fluid surface is reconstructed from the particles in the view space with depth values and surface normals, and then rendered directly in the screen space with per-fragment lighting.
However, in its most simplistic form, this approach does not consider multiphase fluid particles, such as with different materials and colors.
\begin{figure*}[t]
\centering
    \subfloat[\label{fig:fluid_demo_1}]{%
      \includegraphics[width=0.24\textwidth]{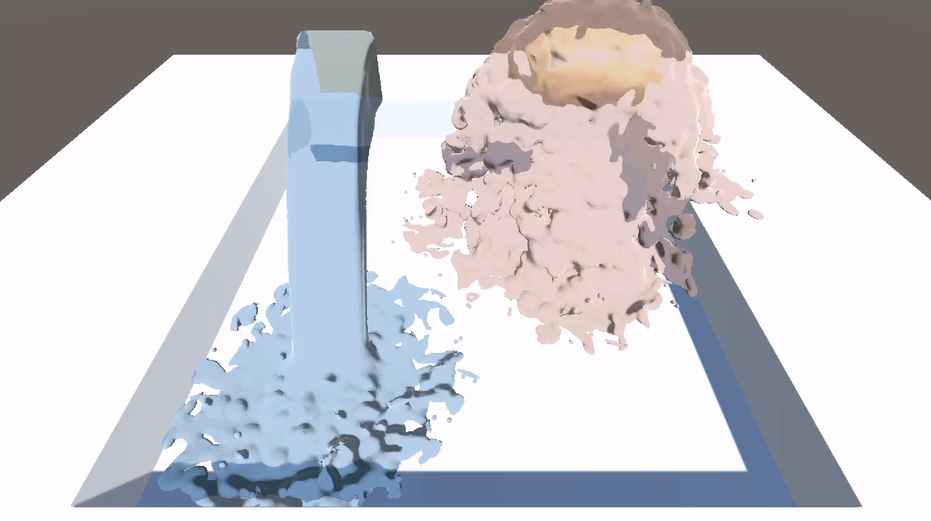}%
    }\hfil
    \subfloat[\label{fig:fluid_demo_2}]{%
      \includegraphics[width=0.24\textwidth]{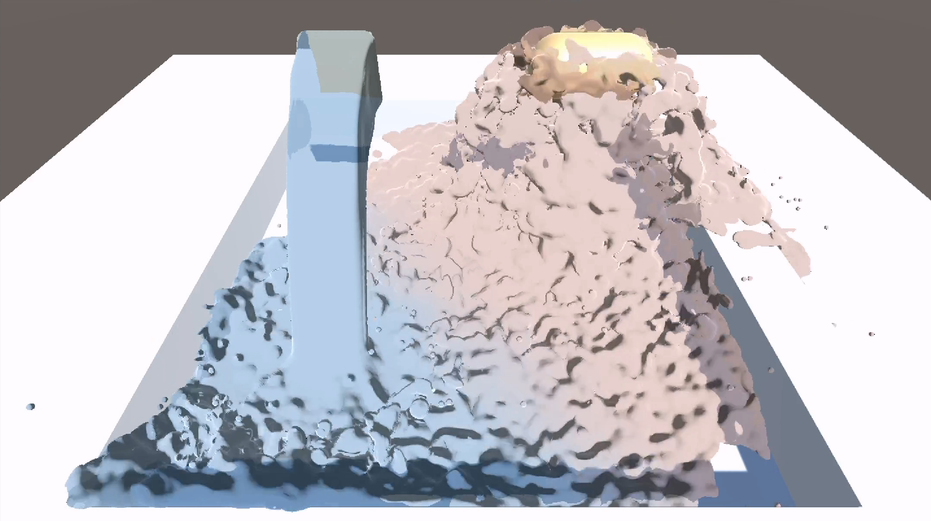}%
    }\hfil
    \subfloat[\label{fig:fluid_demo_3}]{%
      \includegraphics[width=0.24\textwidth]{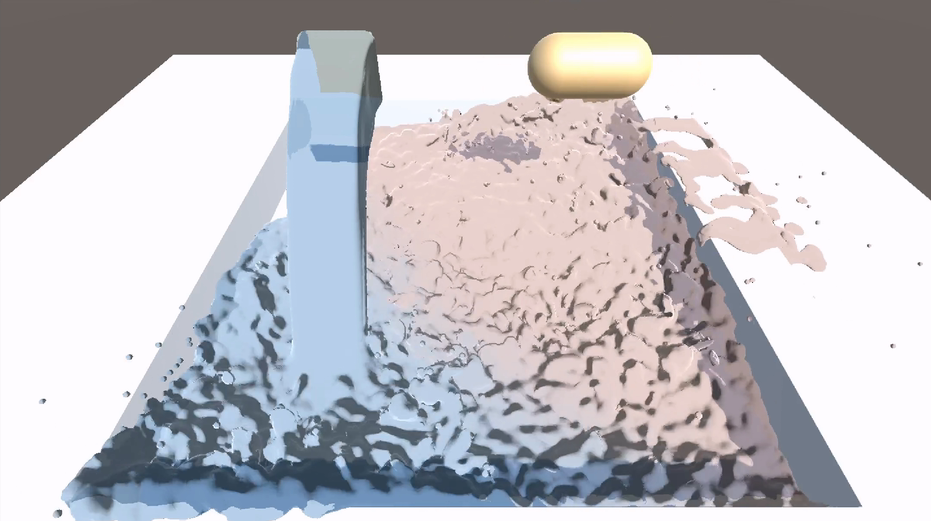}%
    }\hfil
    \subfloat[\label{fig:fluid_demo_4}]{%
      \includegraphics[width=0.24\textwidth]{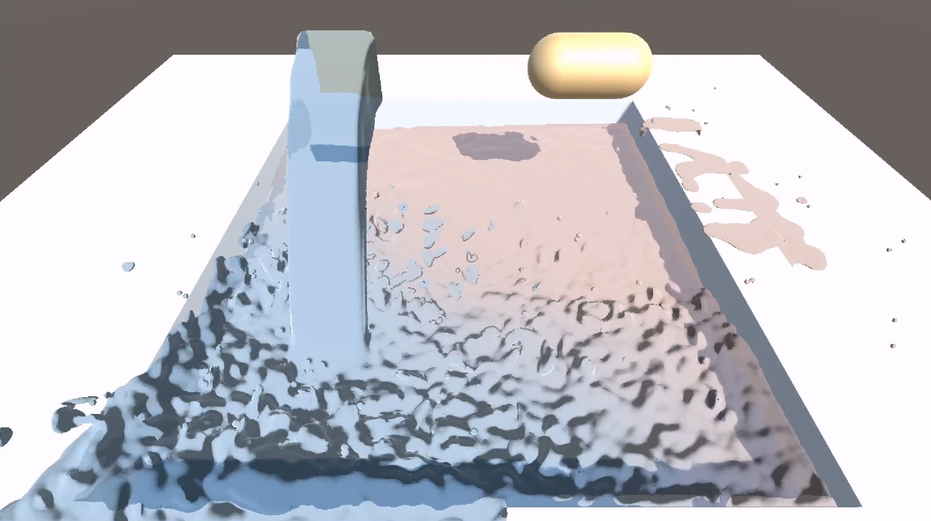}%
    }
A sequence of screenshots from a demonstration scene illustrating the simulation's capability in fluid physics and rendering. The images showcase the interaction between two flowing fluids, their mixing behavior, and visual effects.
\label{fig:fluid_demo}
\end{figure*}
\begin{figure}[t]
    \centering
    \includegraphics[width=0.9\columnwidth]{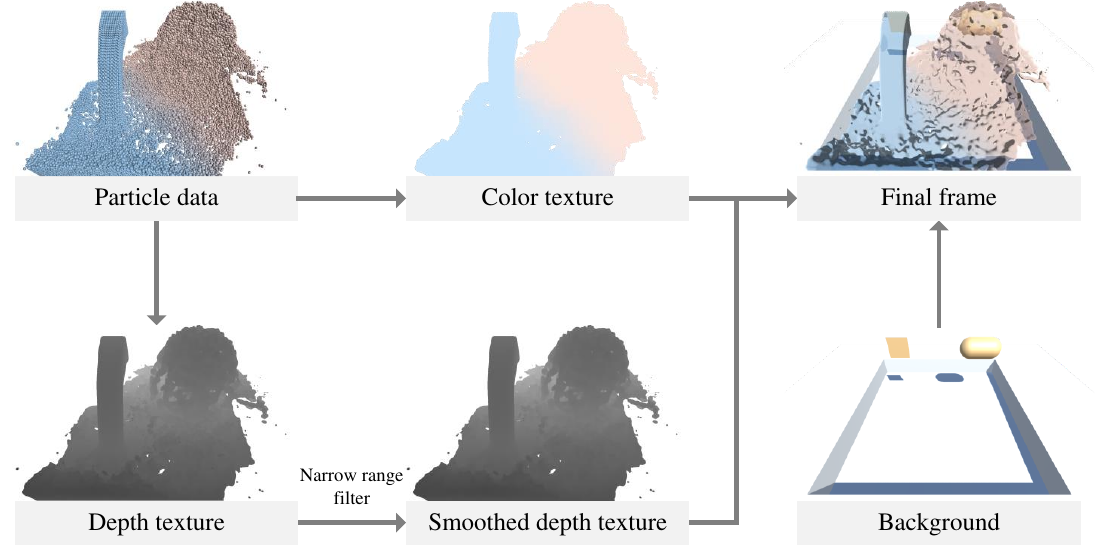}%
    \caption{Fluid surface rendering procedure.}
    \label{fig:render_flow}
\end{figure}

However, rendering should account for the different colors of each particle in our case. To achieve this, an additional color extraction pass is added to the regular screen-space rendering scheme after the surface depth texture is extracted, which simply extracts the colors of particles that are closest to the fluid surface.
This process is similar to \cite{zhang2024real}, where the authors propose using a separate pass to extract the colors based on the thickness of the fluid.
The difference is that we do not consider any particle colors apart from the surface ones. This eliminates the need for a thickness pass to extract the fluid thickness and is a valid simplification without losing much visual plausibility, thanks to the previously discussed color diffusion scheme.

We use anisotropy ellipsoids discussed in \cite{xu2023anisotropic} when reconstructing the fluid surface and a narrow range filter \cite{truong2018narrow} to smooth it.
An example fluid simulation and rendering scene is shown in Fig.~\ref{fig:fluid_demo}.
The overall fluid rendering steps are shown in Fig.~\ref{fig:render_flow}.
In a stress test scene with two large mixing fluid blocks of different colors, each containing 125 thousand particles, the average render time is approximately 7.5 ms per frame (measured over multiple frames).
The physics computation is the primary bottleneck, requiring an average of 24.1 ms per physics step.
Considering task parallelism and all other operations, the frame rate can generally be achieved around 30 Hz.
This test was conducted with an Nvidia RTX 4070, a mid-tier consumer GPU.
As a comparison, a scene with 20 robots takes less than 3 ms for physics computation per step.

\subsection{A Surgical Robot Learning Framework for the dVRK}
\begin{figure}[ht]
    \centering
    \subfloat[]{%
    \includegraphics[height=0.35\columnwidth]{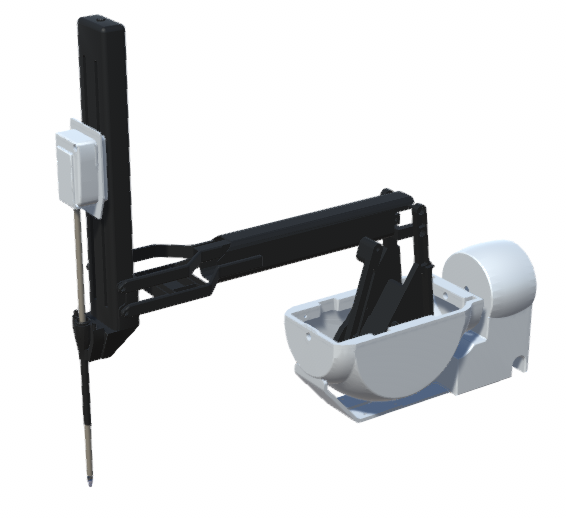}%
        \label{fig:sim_psm}%
    }\hfil
    \subfloat[]{%
    \includegraphics[height=0.35\columnwidth]{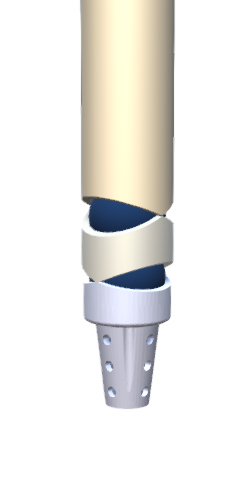}%
        \label{fig:sim_suction_irrigator}%
    }\hfil
    \subfloat[]{%
    \includegraphics[height=0.35\columnwidth]{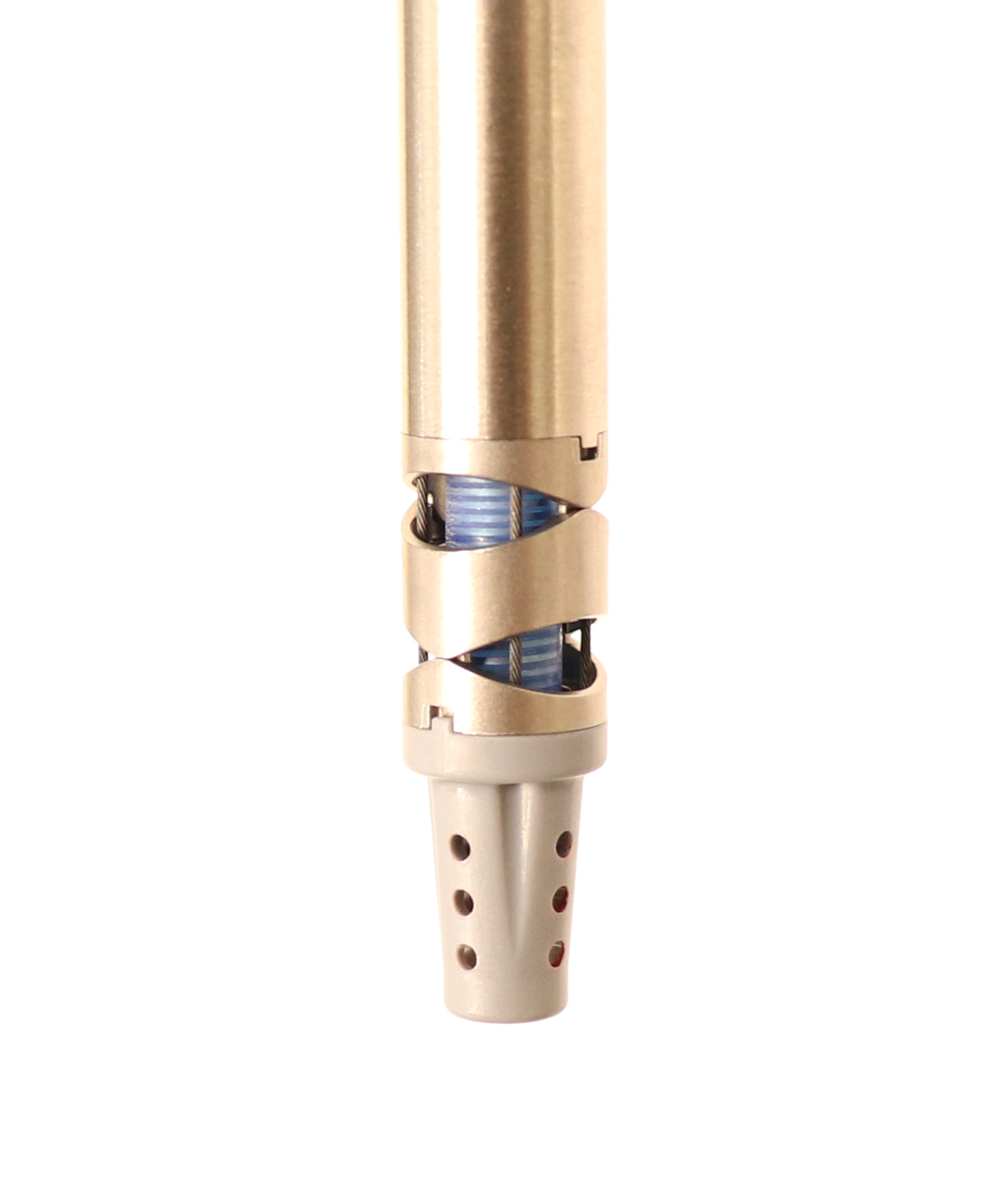}%
        \label{fig:sim_suction_irrigator_real}%
    }
    \caption{(a) Simulated PSM with the EndoWrist One Suction/Irrigator; (b) Simulated Suction/Irrigator tooltip; (c) Real Suction/Irrigator tooltip.}
    \label{fig:simulated_psm}
\end{figure}

In our previous work \cite{ou2024realistic}, a general surgical simulation platform (CRESSim) has been built by leveraging PhysX 5 and Unity, where the dVRK PSM robot is simulated and the real-world dVRK MTM robot is used for teleoperating the simulated PSM.
The PSM is a robot with a mechanical remote center of motion (RCM).
The robot, together with the EndoWrist One Suction/Irrigator, is simulated as a kinematic tree with multiple articulation joints, as shown in Fig.~\ref{fig:simulated_psm}.
It is worth noting that the Suction/Irrigator has only two degrees of freedom (DoF), making the entire PSM a 5-DoF robot.

\begin{figure}[t]
    \centering
    \includegraphics[width=1\columnwidth]{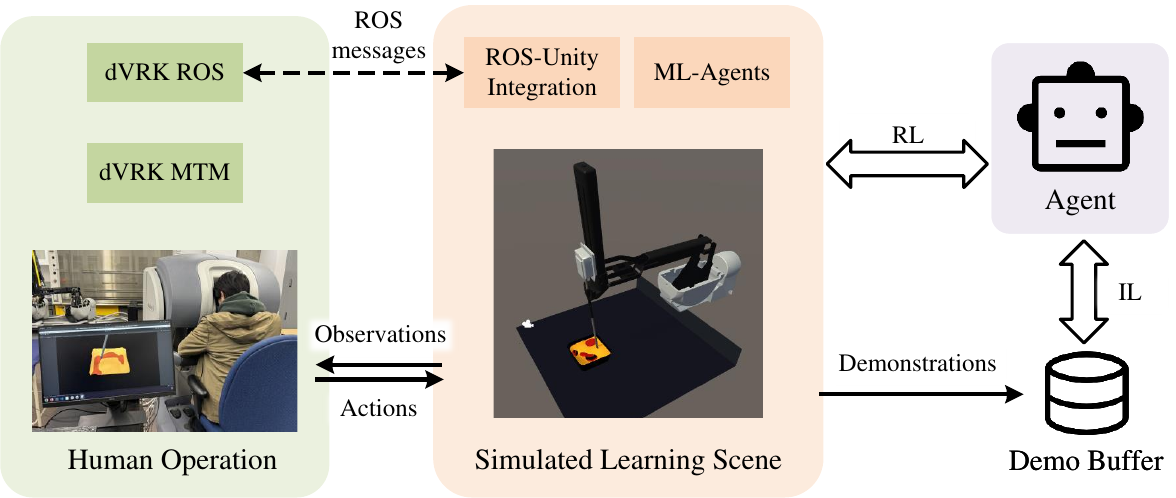}%
    \caption{A surgical robot learning platform for the dVRK with Unity, ML-Agents, and PhysX 5. The human operator sees the simulated camera in the console and teleoperates the simulated robot for collection demonstrations.}
    \label{fig:learning_platform}
\end{figure}

Based on this, we are able to further implement a robot learning framework by incorporating the Unity Machine Learning Agents Toolkit (ML-Agents) \cite{juliani2018unity} into the existing software.
To allow parallel training environments, additional modifications were made for long-term episodic resetting of the environments, such as fluid particle and material property resetting.
Additional features are introduced, such as shared object meshes and shapes between duplicate training areas, which can potentially save memory usage for large-scale training.
Human demonstrations for completing simulated tasks can then be collected while using the real-world MTM to teleoperate the PSM in the simulated learning scene, as shown in Fig~\ref{fig:learning_platform}. The procedures for collecting demonstrations will be further discussed in Section~\ref{sec:lfd}.

Based on this platform, two simulated learning scenes for irrigation and suction are developed, which will be discussed in the following section.
While we only consider irrigation and suction in this work, the platform can be generally used for building various surgical robot learning tasks for the dVRK, including those that use different instruments such as the Large Needle Driver.
Previous work has shown its capability to simulate a number of surgical tasks, such as tissue cutting and manipulation \cite{ou2024realistic}.
As the proposed platform combines CRESSim with ML-Agents, we refer to it as CRESSim-ML.

\subsection{Learning Environments for Irrigation and Suction}
\label{sec:learning_envs}
\begin{figure}[ht]
    \centering
    \includegraphics[width=0.7\columnwidth]{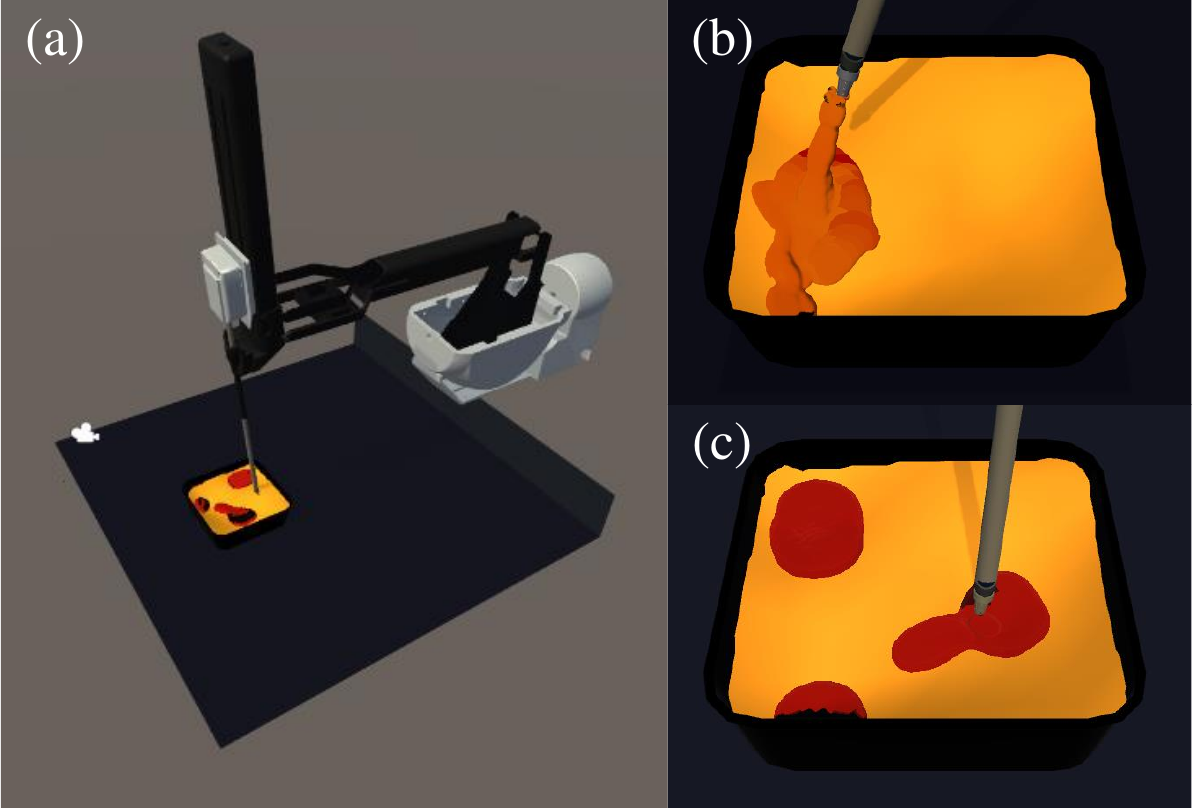}%
    \caption{Simulated training environments. (a) Overview of the scene setup; (b) Irrigation learning environment; (c) Suction learning environment.}
    \label{fig:sim_env_overview}
\end{figure}

Based on CRESSim-ML, two simulated learning scenes are built for training agents to complete the irrigation and suction tasks autonomously and to achieve sim-to-real transfer.
The overall configurations of both scenes are similar, consisting of a PSM robot with the Suction/Irrigator, a tissue container where fluids are present, a camera for generating image observations, and a background tabletop, as shown in Fig.~\ref{fig:sim_env_overview}a.
To ensure the trained policy can be applied to the real-world setup, the simulated scenes are designed to resemble the real world, as will be discussed in Section~\ref{sec:experiments}.
It is worth noting that the environment setup is not intended to realistically represent a specific surgical procedure but rather to validate the automation of the surgical task considered in this work.

In this work, we treat irrigation and suction as two sub-tasks completed by two separate agents, instead of training a unified agent that completes the full process.
A key challenge in training a unified agent is defining an effective reward function.
While the purpose of suction is well-understood, the goal of irrigation is usually ambiguous when described mathematically.
If the reward is based solely on complete contaminant removal, the agent will learn to prioritize suction alone, even though irrigation is beneficial for making the contaminant less sticky.
In fact, without irrigation, it is still possible to suction and remove most contaminants in practice, as long as the Suction/Irrigator is in close contact with the contaminant.
However, irrigation not only aids in dilution but also prevents contaminants from sticking within the Suction/Irrigator.
Therefore, considering two separate sub-tasks is a more natural choice.
Moreover, separating the agents provides practical flexibility.
For instance, in applications like blood suction (as explored in previous studies, \eg{\cite{richter2021autonomous,huang2021model,ou2024autonomous}}), our suction agent could be deployed independently without the irrigation phase.

It is also worth noting that in general, we consider models that observe a vision input from the camera and a vector input of the robot joint, and output the joint-space position increment.
Since the action is in the joint space, the observation of the current joint positions is needed.
This allows direct mapping to low-level controller commands in the joint space without requiring an inverse kinematic solver.
In this work, using Cartesian end-effector (EE) position increments as agent actions is also impractical, as the PSM combined with the Suction/Irrigator has only 5 DoFs.
A Cartesian space action requires 6-DoF position increments, but since the robot is kinematically constrained, it lacks the necessary DoFs to support this approach.
For example, if the agent's output specifies pure translation of the end-effector along the X-axis while keeping all other DoFs at zero, the system will typically introduce an unintended rotational component. This occurs because the robot has only 5 DoFs, making it kinematically impossible to achieve arbitrary 6-DoF Cartesian motion. As a result, the robot moves in a way that satisfies its own constraints, violating the agent's intended zero-motion in other DoFs.
Ideally, there should be no significant performance difference between joint-space and Cartesian actions, provided the current robot position is included in the observation.
Furthermore, we employ incremental actions rather than absolute joint positions, as is common in RL-based control work, such as in \cite{tagliabue2020soft,xu2021surrol,shahkoo2023deep,xian2022fluidlab}.
Incremental actions mitigate training challenges caused by absolute-value scaling, where small workspace variations become difficult to learn.
Additionally, they help maintain consistent execution timing, preventing large, stepwise motions that could disrupt inference-time control consistency.

\subsubsection{Irrigation Learning Environment}
In this task, we consider the problem of diluting condensed contaminants, such as clotted blood, by irrigating liquid from the Suction/Irrigator. While transparent saline is generally used in practice, we consider irrigating a red-colored liquid as a simplification.
This ensures all liquid added into the container is distinguishable from the tissue and can be further suctioned by the suction agent in real-world experiments.
The simplification is reasonable since saline usually turns red instantly in practice after being irrigated due to the presence of blood.

In each episode, a random tissue shape is generated in the container, as will be elaborated in Section~\ref{sec:randomization_and_cl}. A random amount of fluid block with high cohesion and friction and a dark red color that simulates a clotted blood area is dropped onto the tissue.
The goal is for the agent to control the PSM's joint-space motion, move the EE toward the dense blood area, and irrigate with liquid of lighter color and less dense fluid properties.
When both types of liquid mix with each other, colors diffuse between particles as discussed in Section~\ref{sec:fluid_sim_render}. Additionally, each particle is initially assigned a value indicating whether it belongs to the clotted blood or the irrigation liquid. This value is also diffused across particles, providing information on the number of particles affected by irrigation. A threshold is used when determining whether the dense blood particles have been affected.
Particles that spill outside of the tissue container will be removed and not considered, as they are set as inactive.
After enough blood particles are affected, the task is considered completed.
A screenshot from the irrigation environment is shown in Fig.~\ref{fig:sim_env_overview}b.

\textbf{Observations and actions\;}
The observation consists of both visual and vector parts.
The visual component includes an RGB image from the camera at each step, as well as an initial frame captured at the beginning of the episode.
The initial frame provides implicit information about the dense blood's starting location.
This is crucial because, as more liquid is added, it becomes difficult to determine where the dense blood was initially, as the liquid spreads and covers the area.
The vector observation includes current robot joint positions and a number indicating whether the robot is in contact with the tissue, which is decided by measuring the EE link's incoming joint force and torque.
The vector observation is stacked with values from the previous 3 steps to include historical motion data.
The actions are the incremental movements of the robot joints, along with a control signal to toggle the irrigation on and off. The action frequency is 10 Hz.

\textbf{Reward function\;}
The reward function is composed of several parts, including a reward proportional to the number of particles affected by irrigation at each step, a reward when the EE moves horizontally closer to the blood, and a completion reward.
A reward is also given for activating irrigation when the EE is close to the blood, and deactivating it when they are not close. Conversely, it penalizes activating irrigation when distant or deactivating it when near. This encourages the agent to only turn on the irrigation while the EE is close to the blood.
Without these, the agent can easily learn a suboptimal solution of always turning on irrigation, which is not a desired behavior in practice.
Additionally, penalties are given when the EE's orientation deviates from being vertical, and when the EE is in contact with the tissue, to encourage safer behavior of the agent in the real world.
While the penalty related to EE's orientation is always assigned, the joint limits of the robot are implemented and respected according to the real-world robot configuration, and may prevent the EE from achieving a fully vertical orientation in certain configurations.
For clarity, the reward function can be expressed by the summation of the reward features $\psi^{r}_i$ multiplied by their weights $w^{r}_i$:
\begin{equation}
    r = \sum_i \psi^{r}_i w^{r}_i,
\end{equation}
as listed in Table~\ref{tab:reward_fun}.
These weighting values are selected based on the scale of the features, and training trials are conducted to tune them for ideal performance.

\begin{table}[t]
\caption{Reward Function Components.}
\centering
\begin{tabular}{lr}
\hline & \\[-1.5ex]
\multicolumn{1}{c}{\textbf{Reward feature $\psi^{r}_i$}}              & \multicolumn{1}{c}{\textbf{Weight $w^{r}_i$}} \\[1ex] \hline & \\[-1.5ex]
\multicolumn{2}{c}{\textbf{Irrigation}}                                                         \\[0.7ex]
Particles affected by irrigation since last step          & 0.2                                 \\
Task completion                                           & 5                                   \\
Change in EE's horizontal distance to dense blood       & 10                                  \\
Irrigation activated near blood, deactivated when distant & 0.02                                \\
Irrigation deactivated near blood, activated when distant & $-0.05$                               \\
Deviation of EE orientation from being vertical           & $-0.00005$                            \\
EE in contact with the tissue                             & $-0.001$                              \\
\multicolumn{2}{c}{\textbf{Suction}}                                                            \\[0.7ex]
Particles suctioned since last step                       & 0.03                                \\
Task completion                                           & 5                                   \\
Change in EE's horizontal distance to nearest liquid       & 5                                  \\
Deviation of EE orientation from being vertical           & $-0.0001$                             \\
EE in contact with the tissue                             & $-0.03$                               \\ \hline
\end{tabular}
\label{tab:reward_fun}
\end{table}

\subsubsection{Suction Learning Environment}
The suction task considers removing the liquid from the container using the Suction/Irrigator, a step typically following irrigation. In the simulated learning environment, similar configurations to the irrigation environment are used. In each episode, a random number of fluid blocks with various amounts of liquid are dropped from different locations into the tissue container, forming a variety of initial liquid areas that simulate a mixture of blood, saline, and other possible contaminants.
The goal is for the agent to control the PSM's EE to suction all liquid in the container away. The suction model follows the implementation of \cite{ou2024autonomous}, where a cone-shaped force field is applied around the EE of the PSM to move all particles in the region close to the force center.
While it may be appropriate to use the terminal states of the irrigation task as the starting point for the suction task, this approach could limit the range of scenarios the suction agent can handle.
As mentioned earlier, our goal is to develop a general suction agent that is less dependent on the irrigation process.
Particles that are close enough to the force center are removed from the scene.
In this task, suction is always turned on and is not controlled by the agent for simplicity.
A snapshot from the suction environment is shown in Fig.~\ref{fig:sim_env_overview}c.

\textbf{Observations and actions\;}
The observations and actions are very similar to those in the irrigation environment. The observation includes the RGB image from the camera and the stacked vector observation of current robot joint positions, along with a number indicating whether the robot is in contact with the tissue. As the initial camera frame is, in principle, not necessary for making a decision, it is not provided as part of the observation in this task.
The actions are the incremental movements of the robot joints without any additional control signals,  as suction is always turned on.

\textbf{Reward function\;}
Same as in \cite{ou2024autonomous}, the agent is rewarded when particles are removed from the tissue container.
However, it is challenging to train the agent using this reward alone due to the sparsity of the reward, as it is likely that no liquid particles are removed most of the time.
As more liquid is already removed, it is increasingly challenging for the agent to explore successful particle removal due to the few number left.
This issue is signified by the existence of multiple different liquid areas caused by various tissue shapes and multiple random initial liquid blocks.
To overcome this issue, the reward function also includes a reward when the EE is approaching the nearest liquid region.
Task completion reward, EE orientation penalty, and contact penalty are also included, similar to the irrigation task. The reward components are listed in Table~\ref{tab:reward_fun}.

\section{Training in the Simulator}
\label{sec:training}
\subsection{Environment Randomization and Curriculum Learning}
\label{sec:randomization_and_cl}
\begin{figure}[t]
    \centering
    \includegraphics[width=1\columnwidth]{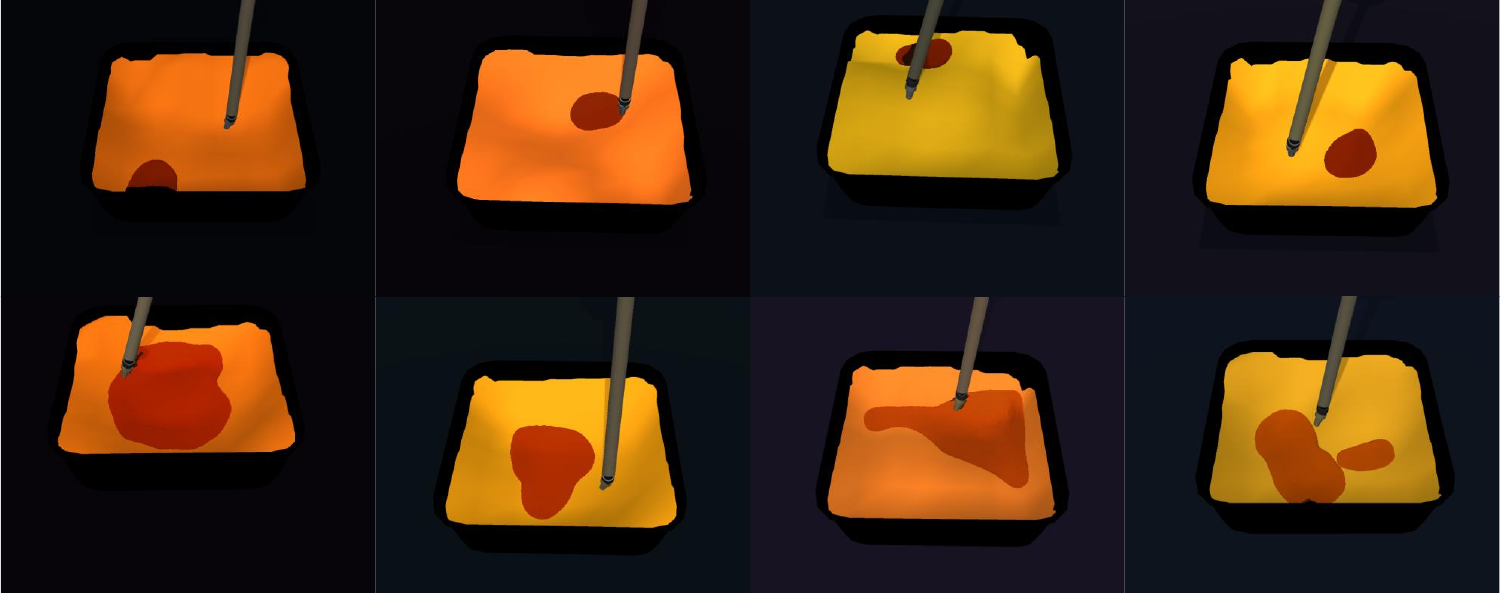}%
    \caption{Examples of training environments with domain randomization for irrigation (upper row) and suction (lower row).}
    \label{fig:random_envs}
\end{figure}

Domain randomization (DR) \cite{tobin2017domain} is a common approach to obtaining a transferrable policy. Similar to \cite{gondokaryono2023learning,haiderbhai2024sim2real,ou2024autonomous}, a number of aspects of the simulation are randomized during training to adapt the agent to diverse visual representations and various types of environments.
Examples of randomized environments are shown in Fig.~\ref{fig:random_envs}.

\textbf{Visual appearance\;}
Visual aspects, including object colors, lighting, and camera pose, are randomized in the simulation.
After setting up the real-world configuration, as will be discussed in Section~\ref{sec:experiments}, camera images are obtained from the real world. Based on the real-world images, the initial colors of various simulated objects are decided.
During training, the object colors are randomly chosen from a range centered on the initial values in the HSV color space.
The objects with randomized colors include the Suction/Irrigator, the tissue in the container, the fluid, and the background tabletop.
Similarly, an estimation of the camera pose in the real world is used to provide an initial value in the simulator, based on which randomness is added to the 3D position and 3 rotational angles during training.
A directional light is used in the simulator, and its pose is randomized in a similar manner. The intensity and shadow strength of the light are also uniformly randomized within a specified range.

\textbf{Task variation\;}
Moreover, we would like the agent to be able to handle a wide variety of task variations, such as with different tissue shapes and different blood locations.
Following previous work \cite{ou2024autonomous}, the tissue shape is randomly generated using a B\'ezier surface by manipulating the control point locations.
The robot's initial joint positions are randomized according to a pre-defined range that roughly places the EE randomly above the container.
For both irrigation and suction tasks, we randomize the initial amount of blood or liquid added to the scene.
As discussed in Section~\ref{sec:learning_envs}, the blood is split into different blocks dropped from different initial locations for the suction task.

\textbf{Physics\;}
Similar to \cite{ou2024autonomous}, we additionally randomize the physics properties, such as the cohesion and viscosity parameters of the irrigation liquid in the irrigation task and the liquid in the suction task. This provides diverse fluid behavior when irrigating and suctioning, and helps bridge the gap between simulation and real-world dynamics.

Curriculum learning (CL) is a common technique to break down a complex task into simpler, more manageable stages and is a well-established method in RL \cite{narvekar2020curriculum}.
We further utilize CL when training the irrigation agent by designing a two-lesson curriculum. In the first lesson, the irrigation reward is turned off, and the agent only learns to approach the blood area without turning on the irrigation, and then turns it on when the EE is close enough to the blood.
Additionally, we do not end the episode when enough particles are affected by irrigation, and no task completion reward is provided.
In the second lesson, we restore the original task rewards to let the agent learn the actual task.
Between these two lessons, we include transitional lessons where both task configurations (with and without the irrigation reward) are randomly sampled according to a proportion. Overall, the proportion of the actual task increases, gradually guiding the agent toward mastering the complete task through CL.
Its effectiveness will be discussed in Section~\ref{sec:training_results}.

\subsection{Learning from Demonstration}
\label{sec:lfd}
We further investigate the effectiveness of leveraging expert demonstrations during training. Two IL methods are used, including behavior cloning (BC) and generative adversarial imitation learning (GAIL).
BC is applied by adding an additional policy loss during training to encourage generating actions close to the demonstrations:
\begin{equation}
    \mathcal{L}_{BC} = \mathbb{E}_{(\mathbf{s}, \mathbf{a}) \sim \mathcal{D}_{E}} \left[ \ell\left( \pi(\mathbf{s}), \, \mathbf{a} \right) \right],
\end{equation}
where $\mathcal{D}_{E}$ is the expert demonstration dataset, $\mathbf{s}$ and $\mathbf{a}$ are the state and action, respectively. $\ell\left( \pi(\mathbf{s}), \, \mathbf{a} \right)$ is a discrepancy measure between the policy-predicted action $\pi(s)$ and the expert action $a$. A mean squared error (MSE) loss is used for continuous actions, and a cross-entropy loss is used for discrete actions.

GAIL is used together with RL to provide additional reward signals that encourage the agent to behave similarly to the demonstrations. A discriminator network $D$ is trained to distinguish expert and agent trajectories during training:
\begin{equation}
\begin{aligned}
    \mathcal{L}_{GAIL} = \: & \mathbb{E}_{(\mathbf{s},\mathbf{a})\sim \mathcal{D}_E} \left[ \log D(\mathbf{s},\mathbf{a}) \right] + \\
    & \mathbb{E}_{(\mathbf{s},\mathbf{a})\sim \mathcal{D}_A} \left[ \log (1 - D(\mathbf{s},\mathbf{a})) \right],
\end{aligned}
\end{equation}
where $\mathcal{D}_A$ is the agent-generated trajectory buffer. The output of the discriminator is then used as an auxiliary reward in addition to the actual environment reward $r^{env}$ with a weighting factor $w$:
\begin{equation}
    r = r^{env} - w \log(1 - D(\mathbf{s},\mathbf{a})).
\end{equation}
A variation of GAIL is used in this study, where only the current observation is used by the discriminator, as it tends to result in more stable training.

Both BC and GAIL are commonly used with RL to take advantage of expert demonstrations and are implemented as part of ML-Agents.
During the initial stage of training, regular RL optimization steps are followed by additional BC steps using $\mathcal{L}_{BC}$ with a separate optimizer.
The learning rate decays linearly to zero, after which no BC steps will be carried out. The auxiliary reward from GAIL is added to the environment reward throughout training.
A similar approach has been previously investigated and shown success on tissue manipulation in \cite{pore2021learning}.

To collect demonstrations with different characteristics and to reduce human effort, two types of demonstrations are collected through both scripted policies and human teleoperation in the simulated environment.
When implementing the scripted policies, the actual particle states are used, although they are not observed by the agent.
For irrigation, all dense blood particles are looped over to find the horizontal center of the blood region, and the target position of the PSM's EE is set to a point higher than that. Irrigation is turned off if they are not close enough, otherwise, it is turned on.
For suction, all particles in the container are clustered into a few regions by building a spatial hash grid, similar to that discussed in Section~\ref{sec:fluid_sim_render}. The EE's target position is set to a point higher than the medoid of the nearest blood cluster.
For both scripted policies, the target orientation of the EE is always vertical. After obtaining the target EE pose, the Jacobian-based iterative inverse kinematics method is used to calculate the current-step joint-space action.

Similarly, when collecting human demonstrations, the pose of the real-world MTM robot's EE is captured and set as the target EE pose of the PSM, from which the per-step joint-space actions are calculated using inverse kinematics. In addition, the MTM gripper pinch signal is used to indicate whether irrigation is on or off for the irrigation task.

For both irrigation and suction tasks, 50,000 steps of demonstration are collected, with half generated from scripted policies and the other half from human teleoperation.

\subsection{Training Configurations}
The physics simulation (environment) step size is 0.02 seconds. However, the decision (action) frequency is lower than the simulation steps at once per 0.1 seconds, resulting in a decision frequency of 10 Hz. In the initial stage of each episode, the PSM robot is driven to a random initial position, and the fluids are dropped from a height into the container, during which no actions will be taken. The maximum number of allowed environment steps is 1,000 and 2,000 for irrigation and suction tasks, respectively. The episode terminates early if the task is considered complete, as described in Section~\ref{sec:learning_envs}.

\begin{figure}[t]
    \centering
    \includegraphics[width=1\columnwidth]{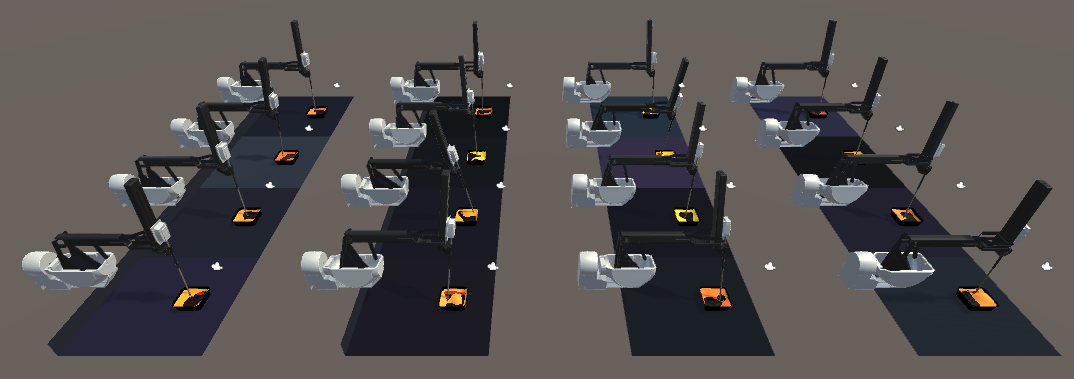}%
    \caption{Unity scene consisting of 16 training areas with random parameters.}
    \label{fig:parallel_training}
\end{figure}

\begin{table}[ht]
\caption{Training Hyperparameters.}
\centering
\begin{tabular}{lr}
\hline & \\[-1.5ex]
\multicolumn{1}{c}{\textbf{Hyperparameter}}  & \multicolumn{1}{c}{\textbf{Value}} \\[1ex]  \hline & \\[-1.5ex]
\multicolumn{2}{c}{\textbf{Common}}                                               \\[0.5ex]
Rollout buffer size                          & 32768                              \\
Batch size                                   & 2048                               \\
Learning rate (linear decay)                 & 3e-4                               \\
Entropy regularization $\beta$               & 1e-2                               \\
Clipping parameter $\epsilon$ (linear decay) & 0.2                                \\
Generalized advantage estimation $\lambda$   & 0.95                               \\
Epochs per update                            & 3                                  \\
Visual encoder type                          & simple (2 layers of CNN)           \\
MLP layers                                   & 3                                  \\
MLP hidden units                             & 128                                \\
\multicolumn{2}{c}{\textbf{Imitation learning (if applicable)}}                   \\[0.2ex]
BC loss strength (linear decay)              & 0.2                                \\
BC steps                                     & 1e4                                \\
GAIL reward strength                         & 5e-2                               \\ \hline
\end{tabular}
\label{tab:training_hyperparams}
\end{table}
Proximal policy optimization (PPO) is used to train agents. Training hyperparameters are listed in Table~\ref{tab:training_hyperparams}. Training utilizes 4 Unity processes, each containing 16 parallel training areas (Fig.~\ref{fig:parallel_training}), totaling 64 parallel training environments. The simulation runs at a higher speed than the actual clock time at a time scale of 2. Training and evaluations are run on a PC with an Intel Core i9-14900K and an Nvidia RTX 4090. The irrigation agent is trained for 10 million environment steps (approximately 7 hours), and the suction agent is trained for 20 million environment steps (approximately 13 hours).

\section{Experimental Setup}
\label{sec:experiments}
\begin{figure}[ht]
    \centering
    \subfloat[]{%
    \includegraphics[height=0.5\columnwidth]{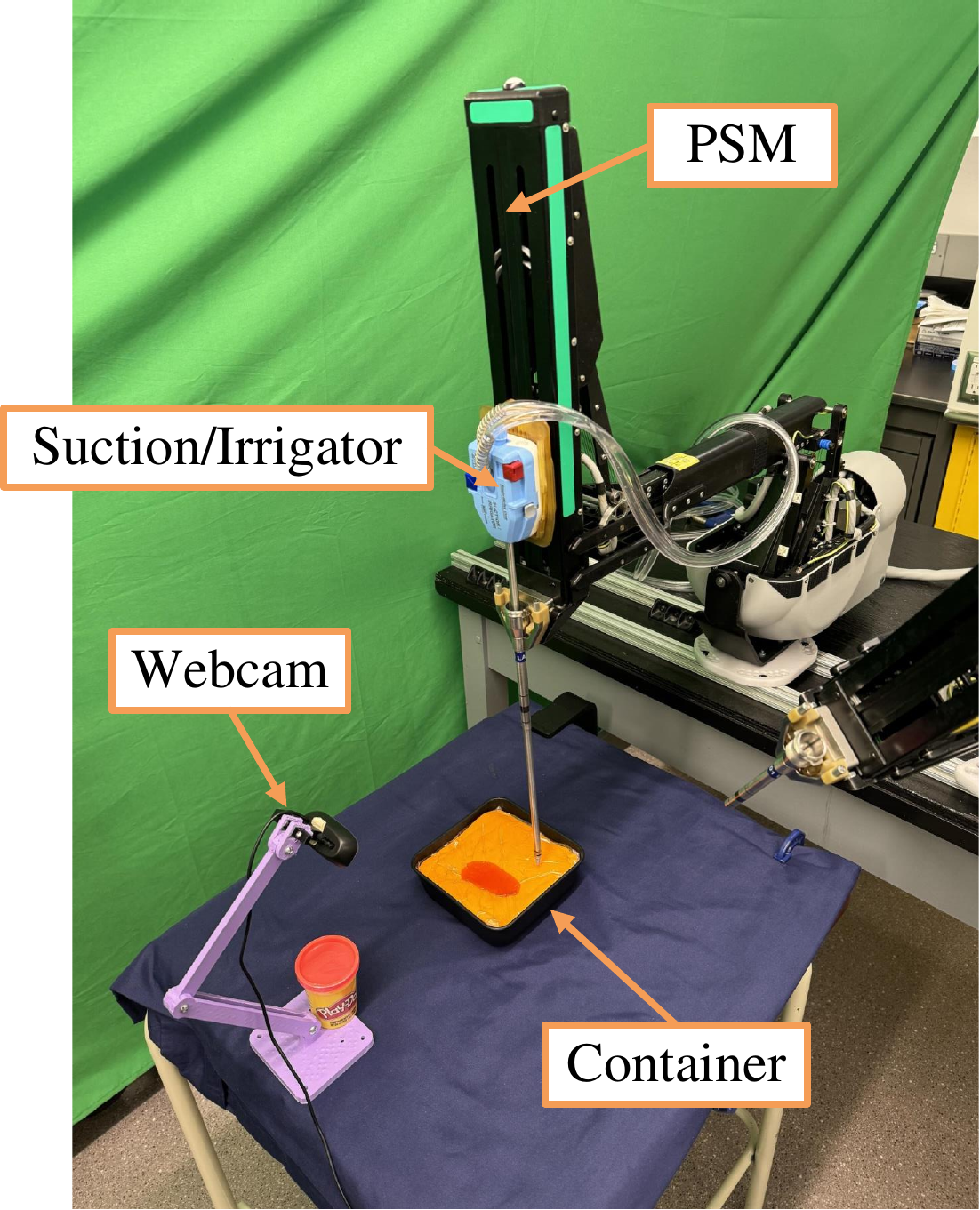}%
        \label{fig:experiment_setup_overview}%
    }\hfil
    \subfloat[]{%
    \includegraphics[height=0.5\columnwidth]{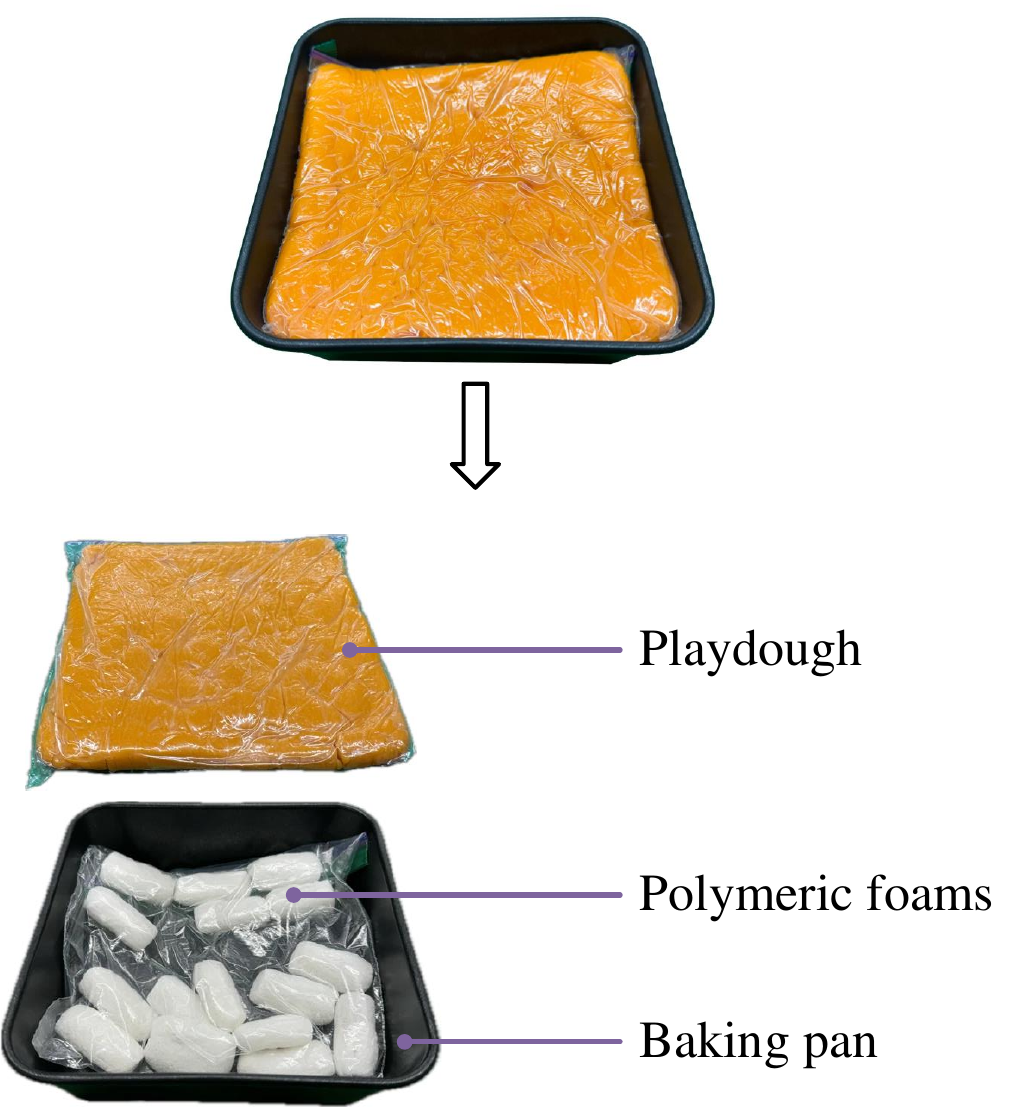}%
        \label{fig:experiment_setup_container}%
    }
    \caption{(a) Real-world setup with the PSM, a webcam, and a container; (b) The container for emulating soft tissue with different shapes.}
    \label{fig:experiment_setup}
\end{figure}

As discussed in Section~\ref{sec:learning_envs}, the real-world setup shown in Fig.~\ref{fig:experiment_setup_overview} is visually similar to the one in the simulator, allowing for direct sim-to-real transfer after training the agents in the simulator. Different tissue shapes can be emulated using a flat piece of playdough placed over various small polymeric foams, as shown in Fig.~\ref{fig:experiment_setup_container}. Changing the foam's positions results in different surface shapes of the playdough.
A baking pan ($15 \times 15$ {cm}$^2$) serves as the base container holding all the components within it.

In the experiments, we first evaluate the performance of the irrigation and suction agents separately on their respective tasks. Then, we assess the overall performance of the complete two-step task, which involves irrigating first and then suctioning. The robot is controlled at around 5 to 10 Hz in the real world, varying due to robot motion execution time at each step.
We intentionally keep the robot motion slower than in training for safety.
The obtained model is converted into the ONNX format with default optimizations by ML-Agents and inferred on Intel(R) Core(TM) i7-9700K.
Image frames are captured from the webcam at each step in an asynchronous manner and fed into the model with the current robot joint state.
The output actions are set as an incremental joint position target and sent to the dVRK software through its ROS package.

\subsection{Irrigation Only}
To evaluate the performance of the irrigation agent, a certain amount (more than 5 grams) of watered-down tomato ketchup is added to the container before running autonomous irrigation.
Since the goal of irrigation is to rinse and dilute the contaminants, and the amount of ketchup being affected by irrigation cannot be measured directly, assessing its performance can be challenging in the real world.
As a workaround, because the irrigation performance ultimately affects how much ketchup can be removed after suction, a manual suction is conducted by the human after each autonomous irrigation trial, assuming that the manual suction is optimal and can remove as much fluid as possible.
We measure the weights before and after the irrigation-suction process, and the change in weight is considered as the irrigation performance. A total of 10 trials are conducted.

\subsection{Suction Only}
To individually evaluate the suction agent, we manually add red-colored fluid that emulates liquid with diluted blood into the container and let the suction agent remove the fluid autonomously. The tissue shape is varied for each trial, with a total of 20 trials conducted.
Of these, 10 trials are initialized with a liquid volume of more than 20 grams, while the other 10 trials begin with more than 30 grams of liquid. Before and after suction, we measure the weight of the fluid in the tissue container to evaluate the amount of fluid removed.

\subsection{Combined Irrigation and Suction}
\label{sec:experiments_combined}
Further experiments are conducted with the agents performing irrigation first and then suctioning autonomously, yielding a complete autonomous irrigation-suction process.
Similar to the irrigation-only experiments, tomato ketchup is added to the container, after which irrigation is performed by the agent autonomously, followed by suction.
To evaluate the overall performance, the weights inside the container are measured before and after the whole procedure.
We further conduct an additional manual suction after the autonomous suction, which clears any remaining fluids that can be suctioned, and record the final weight in the container.
In general, the final weight reflects the irrigation performance, and the difference between the final weight and the weight after autonomous suction reflects the performance of the suction agent.
A total of 10 trials are conducted.

\section{Results}
\label{sec:results}
\subsection{Training Results}
\label{sec:training_results}
\begin{figure}[t]
    \centering
    \subfloat[]{%
    \includegraphics[width=0.7\columnwidth]{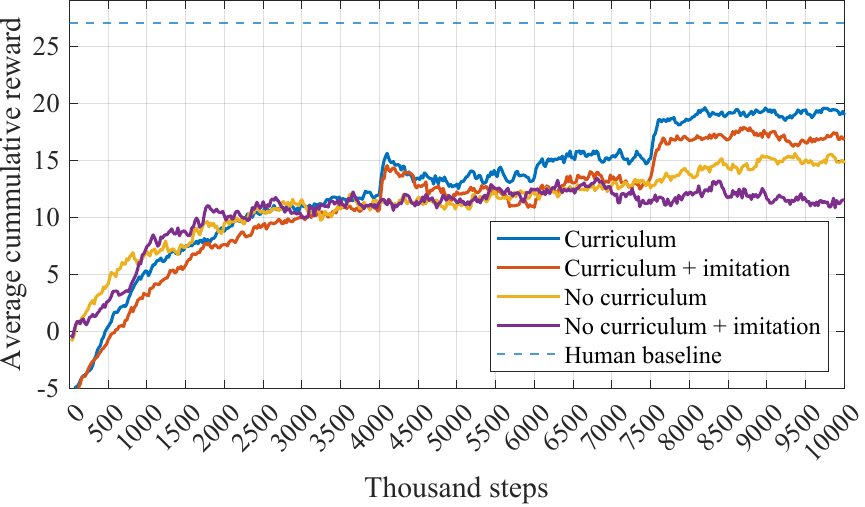}%
    }\\
    \subfloat[]{%
    \includegraphics[width=0.7\columnwidth]{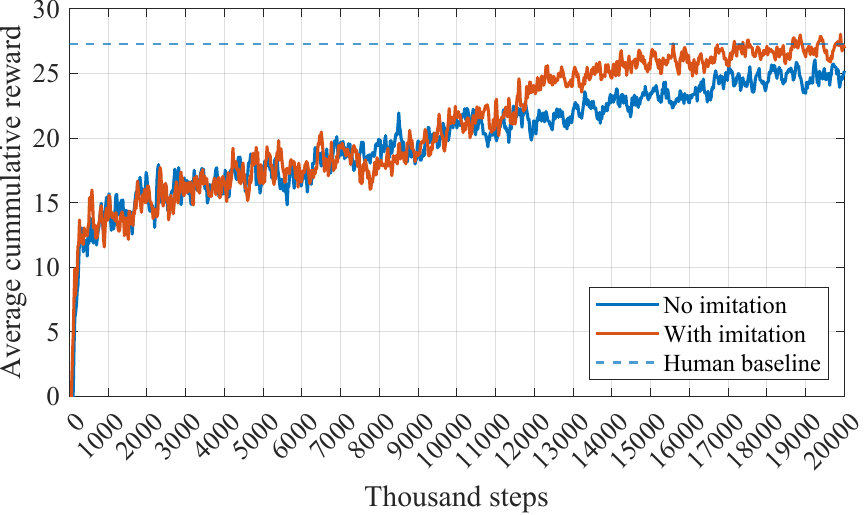}%
    }
    \caption{Learning curves for (a) irrigation and (b) suction. Exponentially smoothed with a window size of 4.}
    \label{fig:learning_curves}
\end{figure}

For irrigation, we conduct training in four distinct configurations: with and without the curriculum, and with and without IL.
This allows us to compare the impact of both CL and IL individually and in combination.
In the case of suction, we only train the agent in two different settings: with and without IL.
Fig.~\ref{fig:learning_curves} shows the learning curves.
It is noticed that for irrigation, the designed curriculum helps achieve better training performance. Conversely, IL does not help the training process for irrigation, in both cases with and without using CL. Instead, adding IL resulted in a counter-effect on training.
However, adding IL resulted in a better learning curve for suction.
The negative effect of IL for irrigation suggests that the reward inferred by GAIL does not fully align with the main RL objective.
This misalignment may arise because the learned GAIL reward signal may not fully guide toward the task-specific objective, or because the agent model cannot capture all of the required features, potentially leading to different optimization directions.

We select the two agents with the highest training return for irrigation and suction, respectively, and evaluate their final performance in the simulator.
A total of 100 trials are conducted for both tasks.
For irrigation, the average return during evaluation is 23.90, and the completion rate is 65\%.
Numerous particles can still be affected by irrigation even if the task is not fully complete.
For suction, the average return during evaluation is 26.24, and the completion rate is 85\%. On average, around 5.52 particles remain in the container after suction. This number is much higher in a few failure cases, reaching more than 100.

\subsection{Real-World Performance of Irrigation Only}
\begin{figure}[t]
    \centering
    \subfloat[]{%
    \includegraphics[height=0.45\columnwidth]{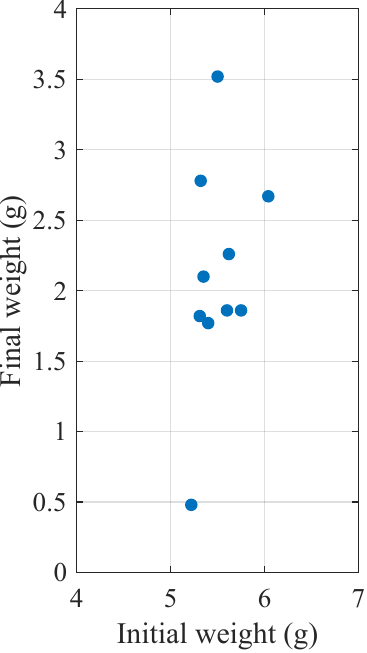}%
    }\hfil
    \subfloat[]{%
    \includegraphics[height=0.45\columnwidth]{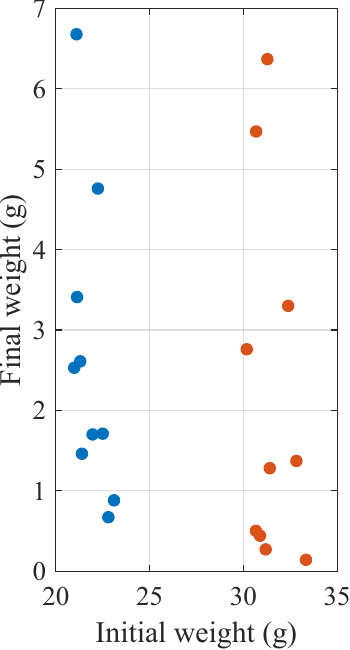}%
    }
    \caption{Results from real-world trials for (a) irrigation only and (b) suction only. The two sets of suction trials are grouped by color.}
    \label{fig:real_world_individual}
\end{figure}

The recorded data for the real-world irrigation trials are plotted in Fig.~\ref{fig:real_world_individual}a. Most of the trials are completed within 5 seconds. On average, 5.53 grams of tomato ketchup is added before irrigation, and after autonomous irrigation and manual suction, the average remaining weight is 2.11 ($\pm \, 0.80$)\footnote{Standard deviation} grams.
As a comparison, manual irrigation by a human operator followed by the same suction step results in an average remaining weight of 1.90 ($\pm \, 0.49$), indicating that the agent's performance is close to that of a human.
Although irrigation is intended to dilute all ketchup in most cases, this is not always possible and typically more than 1 gram of ketchup will remain even in the best-case scenario.
In contrast, if irrigation is not effective and the EE does not aim at the ketchup, much less ketchup will be diluted, leading to a high amount of ketchup remaining.
In reality, surgeons may do irrigation and suction multiple times to ensure that the surgical field is completely cleaned.
However, we only consider the performance of a single irrigation and suction cycle here for simplicity.

\begin{figure*}[t]
    \centering
    \includegraphics[width=0.8\textwidth]{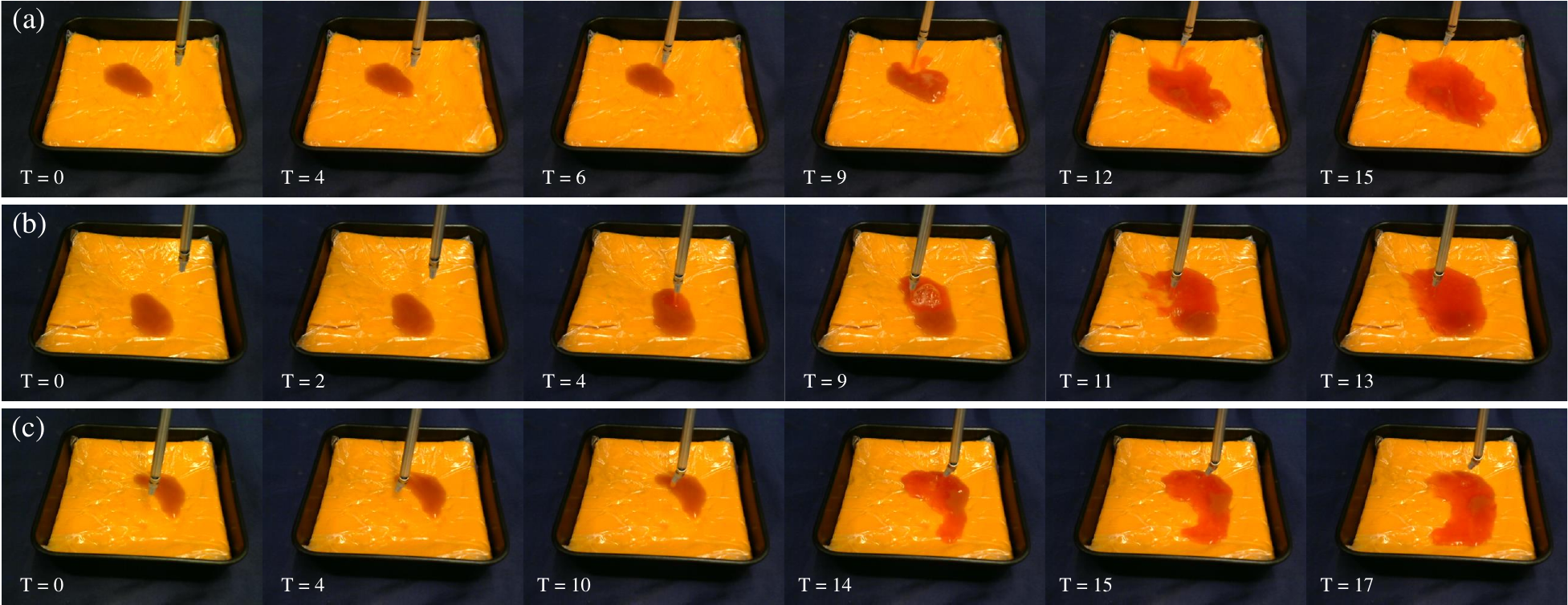}%
    \caption{Representative snapshots selected at different action steps from real-world irrigation experiments.}
    \label{fig:irrigation_snapshots}
\end{figure*}

We are not able to directly compare the real-world performance with the evaluation results in the simulator, as we cannot measure the true amount of ketchup being affected by irrigation.
However, it appears that the real-world performance is slightly inferior to the one in the simulator in terms of accurately targeting the EE toward the ketchup before irrigation.
Snapshots from autonomous irrigation are shown in Fig.~\ref{fig:irrigation_snapshots}.

\begin{figure}[t]
    \centering
    \subfloat[]{%
    \includegraphics[height=0.4\columnwidth]{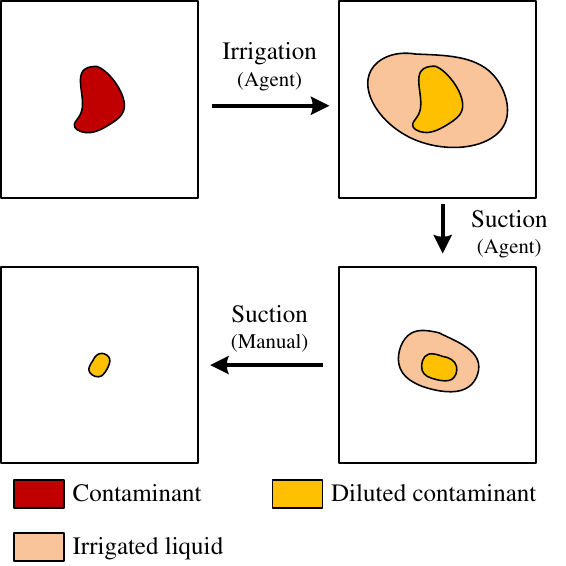}%
    \label{fig:measurement_flow}
    }\hfil
    \subfloat[]{%
    \includegraphics[height=0.4\columnwidth]{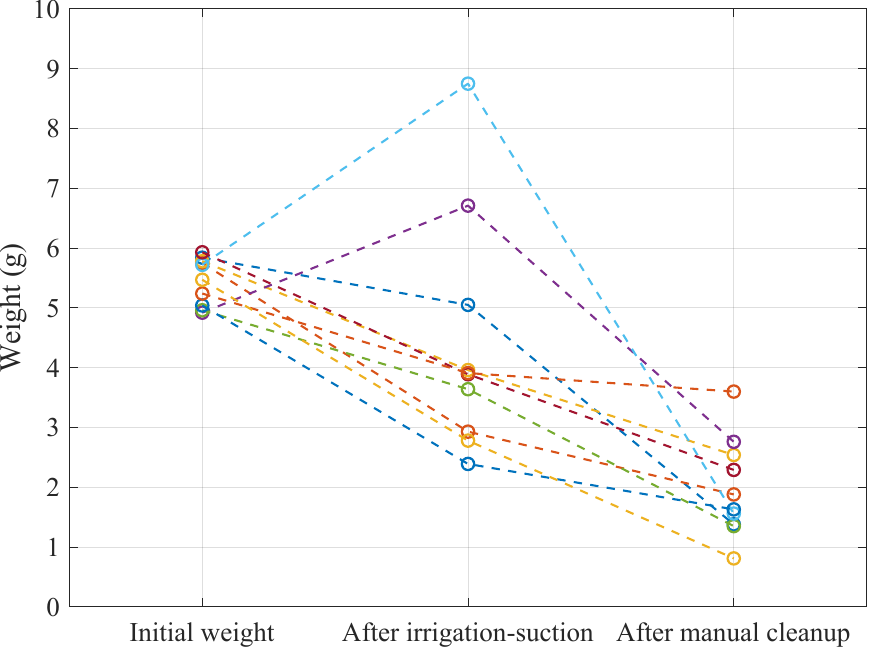}%
    \label{fig:irrigation_suction_results}
    }
    \caption{(a) Experimental procedure and (b) results for real-world combined irrigation-suction. Lines of different colors in (b) represent individual trials.}
\end{figure}

\begin{figure*}[t]
    \centering
    \includegraphics[width=0.8\textwidth]{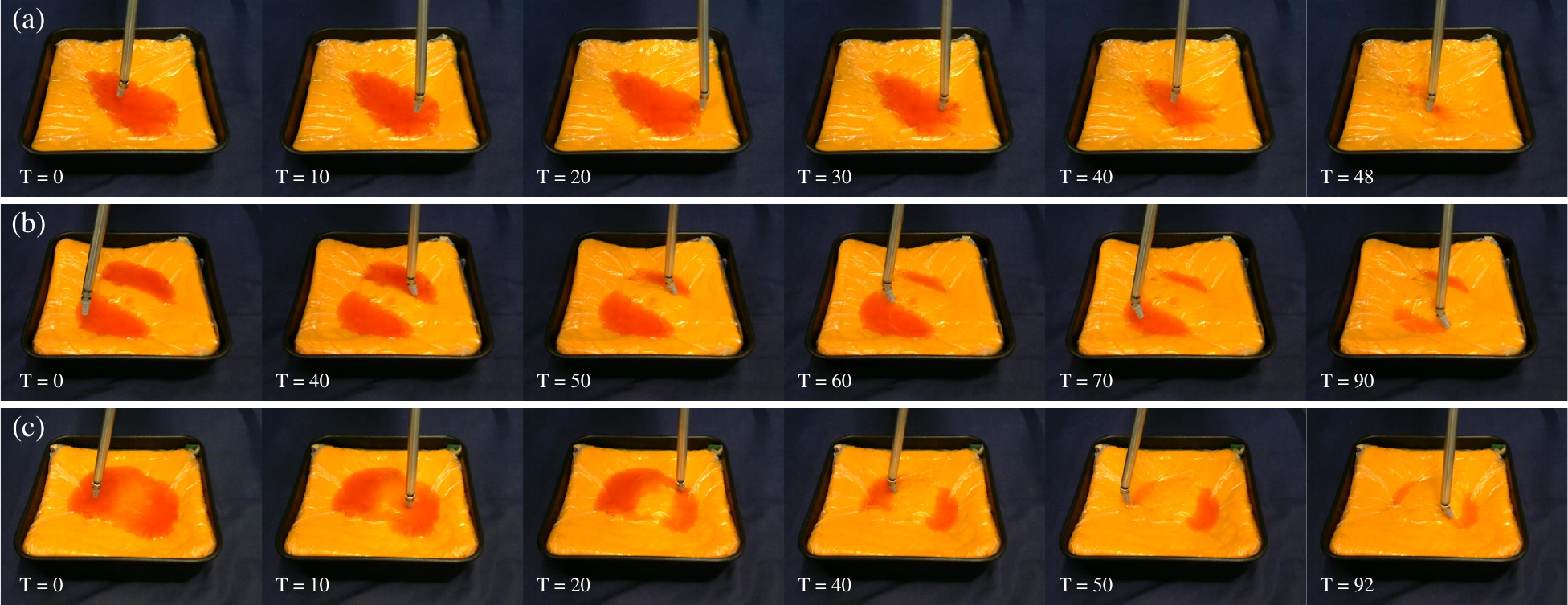}%
    \caption{Representative snapshots selected at different action steps from real-world suction experiments.}
    \label{fig:suction_snapshots}
\end{figure*}

\begin{figure*}[t]
    \centering
    \includegraphics[width=0.8\textwidth]{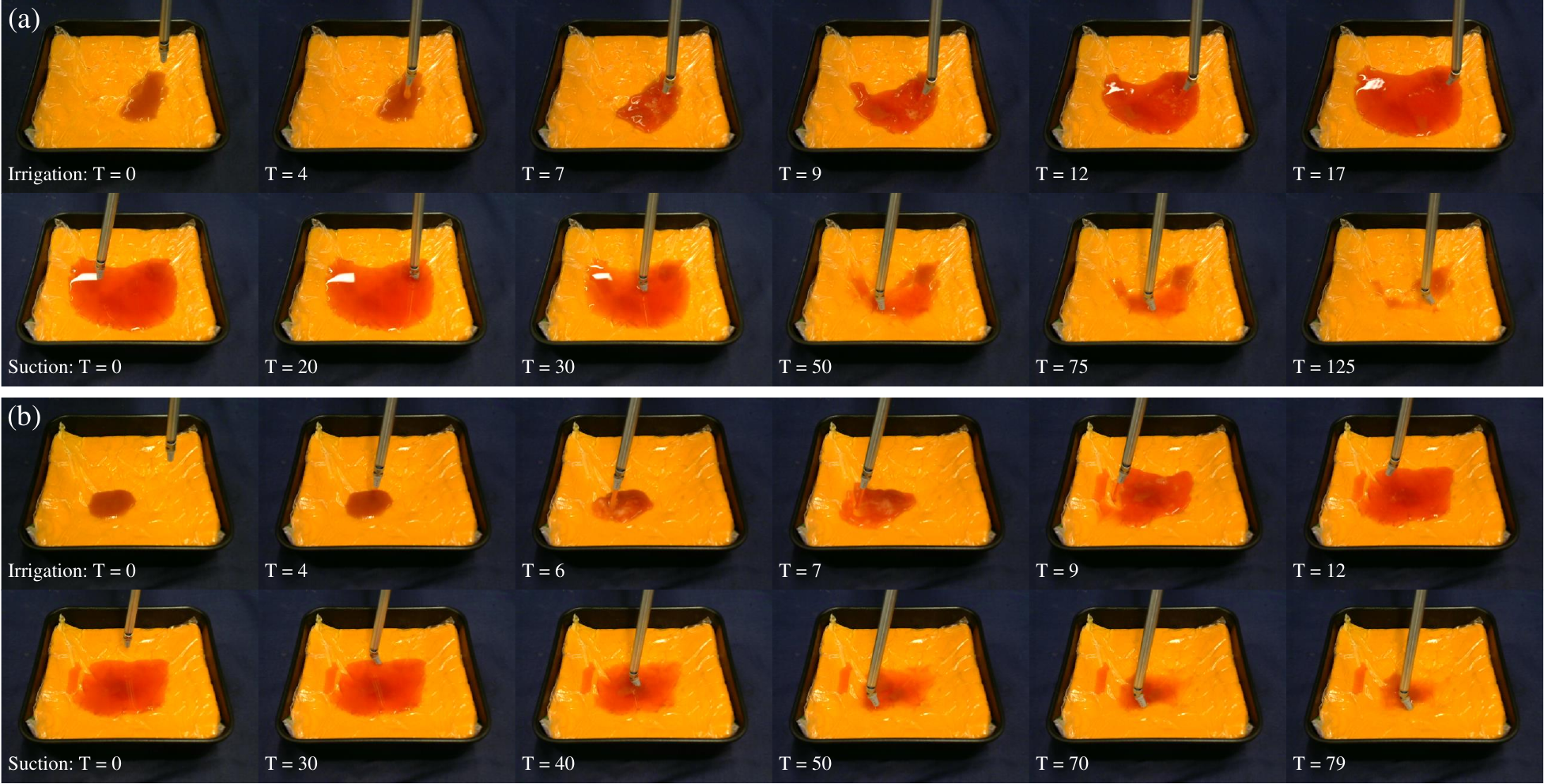}%
    \caption{Representative snapshots selected at different action steps from complete irrigation-suction experiments in the real world.}
    \label{fig:irrigation_suction_snapshots}
\end{figure*}

\subsection{Real-World Performance of Suction Only}
The recorded data for the real-world suction trials are shown in Fig.~\ref{fig:real_world_individual}b. Most of the trials are completed within 30 seconds. The two groups of trials are represented by blue and orange colors on the plot. The average initial fluid weights are 21.85 grams for the first group and 31.49 grams for the second. After suctioning, the average final weights are reduced to 2.64 ($\pm \, 1.87$) grams and 2.24 ($\pm \, 2.24$) grams for the two groups, respectively.
We observe a similar distribution for the final weights across the two groups, suggesting that the agent works consistently for different amounts of initial fluid.
However, within each group, there are trials where a substantial amount of fluid remains after suctioning, which is barely seen when evaluating the agent in the simulator.
This is most likely due to the sim-to-real gap, as will be discussed in Section~\ref{sec:sim_to_real_gap}.
Snapshots from suction-only experiments are shown in Fig.~\ref{fig:suction_snapshots}.

\subsection{Real-World Performance of Combined Irrigation-Suction}
As discussed in Section~\ref{sec:experiments_combined}, during the complete autonomous irrigation-suction experiments, the weight of the contents inside the container is measured three times: the initial weight, the weight after autonomous irrigation and suction, and the weight after an additional manual suction for cleanup, as shown in Fig.~\ref{fig:measurement_flow}.
The results are shown in Fig.~\ref{fig:irrigation_suction_results}.
Snapshots from the experiments are shown in Fig.~\ref{fig:irrigation_suction_snapshots}.

On average, the final remaining weight after manual cleanup is 1.98 ($\pm \, 0.82$) grams on average. Since all liquid is removed at this stage, this value reflects the irrigation performance similar to the irrigation-only trials. The result is consistent with the one from the irrigation-only trials. However, it can be noticed that the actual remaining weight after autonomous irrigation-suction results is higher (4.40 $\pm \, 1.97$ grams) due to residual liquid not being suctioned.

We further calculate the difference between the weight after autonomous irrigation-suction and the one after manual cleanup, resulting in an average value of 2.42 ($\pm \, 2.04$). This value represents the amount of liquid that can be suctioned but is not actually removed during autonomous suction, reflecting the autonomous suction performance, which is similar to the results obtained from the suction-only trials.
An ANOVA test does not find a significant statistical difference between the three groups (combined suction-irrigation, suction-only with around 20 grams of initial weight, and suction-only with around 30 grams of initial weight), with a $p$-value of 0.89.
Since the $p$-value is high, there is insufficient evidence to suggest that suction performance differs across these scenarios, implying that it may be relatively stable.
There is also no strong statistical evidence of correlation between the irrigation and the suction performances, based on the Pearson correlation coefficient and Spearman's $\rho$ ($-0.2953$ and $-0.3939$) and their $p$-values ($0.4075$ and $0.2629$), between (1) the final weight after manual cleanup and (2) the weight difference before and after manual cleanup.
However, we observe more back-and-forth motion during suction this time, compared to the suction-only trials, likely due to the presence of the dark-colored ketchup that is not fully diluted.

\subsection{Suboptimal Outcomes}
\begin{figure}[t]
    \centering
    \includegraphics[width=0.8\columnwidth]{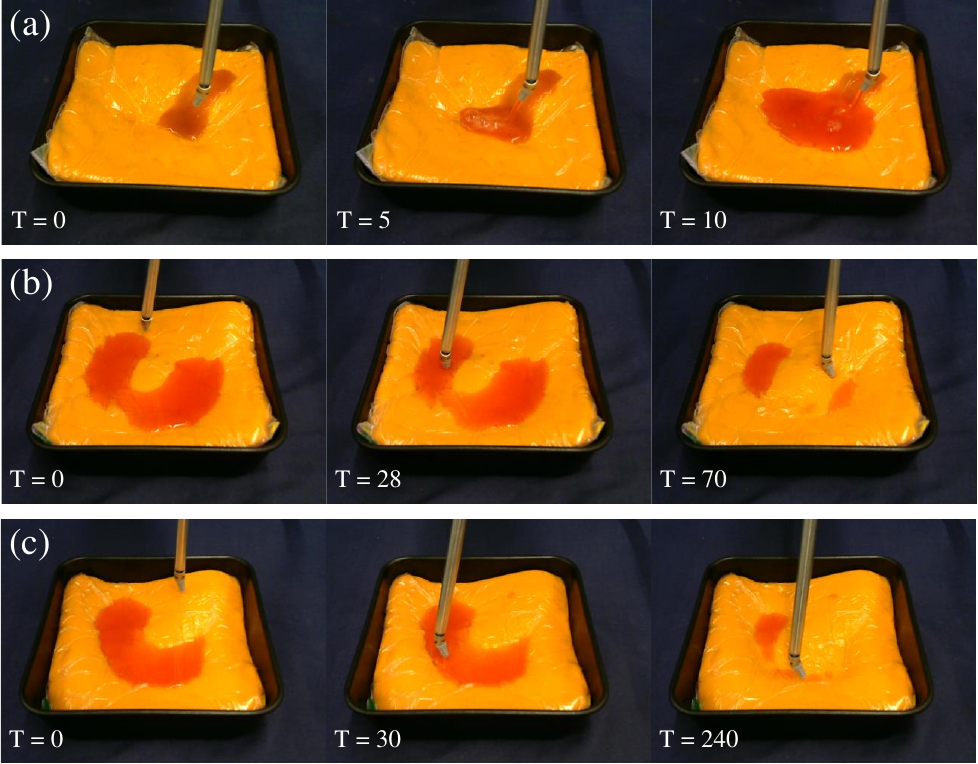}%
    \caption{Suboptimal outcomes in the real-world trials.}
    \label{fig:suboptimal_or_failure}
\end{figure}

A typical suboptimal outcome during irrigation is that the robot's EE does not always aim at the ketchup, leading to some parts of the ketchup not being affected by the irrigation. One example is shown in Fig.~\ref{fig:suboptimal_or_failure}a. In this trial, the EE correctly aimed for the ketchup region initially but failed to continue to target the upper part after irrigation started, leaving the part not irrigated.

For suction, the typical failure case is that the agent fails to navigate the EE to a pool of liquid, which is more likely to occur when the pool is small. In the example shown in Fig.~\ref{fig:suboptimal_or_failure}b, the EE was navigated to the top-left side initially but continued to move to the lower-right before removing all liquid on the top-left side. The agent eventually attempted to move the EE back to the top-left but failed due to the EE touching the container and violating the force constraint, as will be discussed in Section~\ref{sec:other_limitations_future_work}.
In the trial shown in Fig.~\ref{fig:suboptimal_or_failure}c, no attempt was even made to guide the EE to the remaining liquid pool. This is barely seen during the evaluation in the simulator, but is more frequent in the real world.
One possible reason is the sim-to-real gap, as discussed in Section~\ref{sec:sim_to_real_gap}.

\section{Discussion}
\label{sec:discussion}
\subsection{Sim-to-Real Gap}
\label{sec:sim_to_real_gap}
\begin{figure}[ht]
    \centering
    \subfloat[]{%
    \includegraphics[width=0.35\columnwidth]{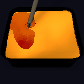}%
    }\hfil
    \subfloat[]{%
    \includegraphics[width=0.35\columnwidth]{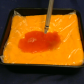}%
    }
    \caption{Image observations for irrigation from (a) the simulated environment and (b) the real world.}
    \label{fig:image_obs}
\end{figure}

There are various discrepancies between the real-world environment and the simulated one.
Since the RGB image is used in this work, one major discrepancy is the difference between the real-world image and the rendered image in the simulator.
Fig.~\ref{fig:image_obs} shows a rendered image from the simulator and one from a real-world trial, in the actual observation size (84 $\times$ 84).
Although the colors and the camera pose are randomized during training, there are uncaptured discrepancies between simulation and reality, such as bright reflections.
Additionally, the liquid color inside the tissue may not be uniform after irrigation, while our suction training environment starts with uniform liquid color in each episode due to implementation simplicity.
The simulated fluid color mixing effect may also not fully emulate that in the real world.
These factors can cause agents to perform worse in the real world compared to their simulation performance.
Certain factors, like reflections, could be better simulated and considered during training to obtain agents more capable of handling or inferring such information.
Otherwise, techniques such as image-to-image translation and domain adaptation (DA) could be explored in future training.

The difference in physics can also enlarge the sim-to-real gap. For example, suction is simulated by a cone-shaped force field, whereas it is achieved primarily by pressure differences in reality. In practice, the EE has to be in contact with the liquid surface to suction it, while in the simulator, particles can be pulled toward the EE even if there is a distance between them.
In addition, a rigid body is used to simulate the tissue, whereas the tissue is emulated using playdough and is soft in reality. The choice of using a rigid body is related to the computational performance considerations, as will be discussed in the following section.

We did not perform a quantitative analysis of sim-to-real transfer performance, including experiments on how well the model generalizes to different camera poses, lighting conditions, and object colors in the real world.
Additionally, control frequency and latency, while not strictly part of the sim-to-real gap, can further impact real-world performance.
To better understand the sim-to-real gap, future work could include further experiments and analysis comparing real-world performance under various conditions with that observed in the simulator.

\subsection{Computational Performance Considerations}
There are several considerations of computational performance.
To allow efficient simulation, a soft body was not used to simulate the tissue although PhysX 5 allows soft body simulation using the finite element method (FEM).
FEM is computationally intensive and requires more GPU resources and processing time.
We could not create a large number of parallel training environments when using FEM soft bodies, nor were we able to achieve efficient training in terms of wall-clock time.
Instead, a rigid body is used to simulate the tissue as a trade-off.
However, this simplification also results in the exclusion of key tissue characteristics such as viscoelastic behavior, especially when the tool is in contact with the tissue.
Liquid permeability, absorption, and surface tension effects are not considered or limited by the contact model between the rigid body and the liquid provided by PhysX 5.
The lack of these properties could reduce the realism of the simulations, potentially affecting the transferability of trained models to real-world scenarios.
While our approach prioritizes computational feasibility and training efficiency, future work may explore other efficient soft body simulation techniques, such as the material point method (MPM) \cite{ou2025cressim} to mitigate these limitations.

We have also considered the observation of historical images to capture the temporal information.
Including historical images can potentially enhance the performance, such as by allowing the agent to infer liquid occluded in the current step by the Suction/Irrigator from previous non-occluded frames.
However, considering the large rollout buffer size and the usage of demonstration data, a much larger amount of memory would be needed.
In the current setting, the memory usage is already more than 30 gigabytes with IL for irrigation, making it impractical to stack the image observations with historical ones.
To solve this issue, either a much larger memory should be used, or ML-Agents must be modified to allow streaming of the rollout data.
Using historical frames at a lower resolution may also be beneficial, though further downscaling from the current $84 \times 84$ resolution could result in significant information loss.

\subsection{Other Limitations and Future Work}
\label{sec:other_limitations_future_work}
In this work, no depth information is provided to the agent. Ideally, it may be possible to achieve a better performance when depth is taken into account.
However, it is not included in this work as it may increase the sim-to-real gap, due to the additional noises and inaccuracies of the real-world depth images.
Future work may either include the depth observation directly or train the agent with a stereo camera that captures two images side-by-side.
Using a stereo camera can also provide additional information that a single camera might miss, in addition to depth, especially when tools occlude the liquid and tissue. However, in this work, significant tool occlusion is not encountered due to the tool’s relatively small radius and the side-view camera pose.

Other imitation learning approaches, such as diffusion policy \cite{chi2023diffusionpolicy} and action chunking with transformers \cite{zhao2023learning,kim2024surgical}, can be further incorporated in future work.
This may allow more in-depth investigation of pure IL approaches without RL, as well as novel approaches that combine the state-of-the-art IL methods with RL.

One limitation of the real-world experiments is that a hardcoded EE force constraint is imposed.
If the force limit is reached when the agent attempts to move the EE toward a position, such as when the EE touches the tissue, the trial is terminated for safety.
This happens only during suction, as close contact between the tool and tissue may be needed to suction the liquid.
Among all the suction trials, including those within combined irrigation-suction, a total of 9 trials were terminated due to force violation.
Although the agent is penalized when the EE is in contact with the tissue during training, force limit violations are still seen during some suction trials.
Future work may employ a better training strategy, such as rewarding the agent for lifting up the EE when it is in contact with the tissue.

This work focuses primarily on obtaining agents that can perform the tasks autonomously under various physical configurations.
While the evaluation focuses on task completion, other performance metrics are less considered, such as the execution time and the frequency of safety violations.
Qualitative observations indicate that irrigation execution time remains stable across trials, whereas suction time varies depending on the complexity of the suction area.
Further analysis of task failure cases and a comparison between agents' performance and human operation may be necessary before considering surgical applications, which are beyond the scope of this research.

Further extension of this work to actual surgeries will also need additional development of realistic surgical scenes and re-training of the agents, as this work considers a relatively simple environment setup without actual surgical components, such as phantom tissue.
With the simulation capability introduced in this work, transitioning to more realistic surgical scenes--including those beyond minimally invasive surgery--is straightforward, particularly when simulating body fluids.
Additional validations are required as well to assess the performance in realistic environments, such as using cadavers.

\section{Conclusion}
\label{sec:conclusion}
In this work, we examined the problem of autonomous irrigation and suction in MIS using RL.
To allow sim-to-real transfer, a new surgical robot learning simulation framework was built using Unity and PhysX 5, which enables the development of simulated learning environments for irrigation and suction.
With DR and carefully designed reward functions, combined with CL and IL, two vision-based agents were trained to autonomously complete irrigation and suction.
We found that the irrigation agent achieves a much better performance when a curriculum is used, and that IL helps improve the suction agent but not the irrigation one.
We evaluated their performance in the real world and showed that they can achieve satisfactory results in most cases.
Additionally, our simulation platform can be extended and used to build simulated training environments for other surgical tasks in the future.

\section{Ackowledgments}
The authors thankfully acknowledge Industry Sandbox \& AI Computing (ISAIC) for providing us with free trials of HPC instances to prototype and test our methods.
They would also like to thank Sadra Zargarzadeh for his support and discussions, and Amir Zakerimanesh and Tleukhan Mussin for their help with the experiments.

\bibliography{references.bib}
\bibliographystyle{IEEETran}

\begin{IEEEbiography}[{\includegraphics[width=1in,height=1.25in,clip,keepaspectratio]{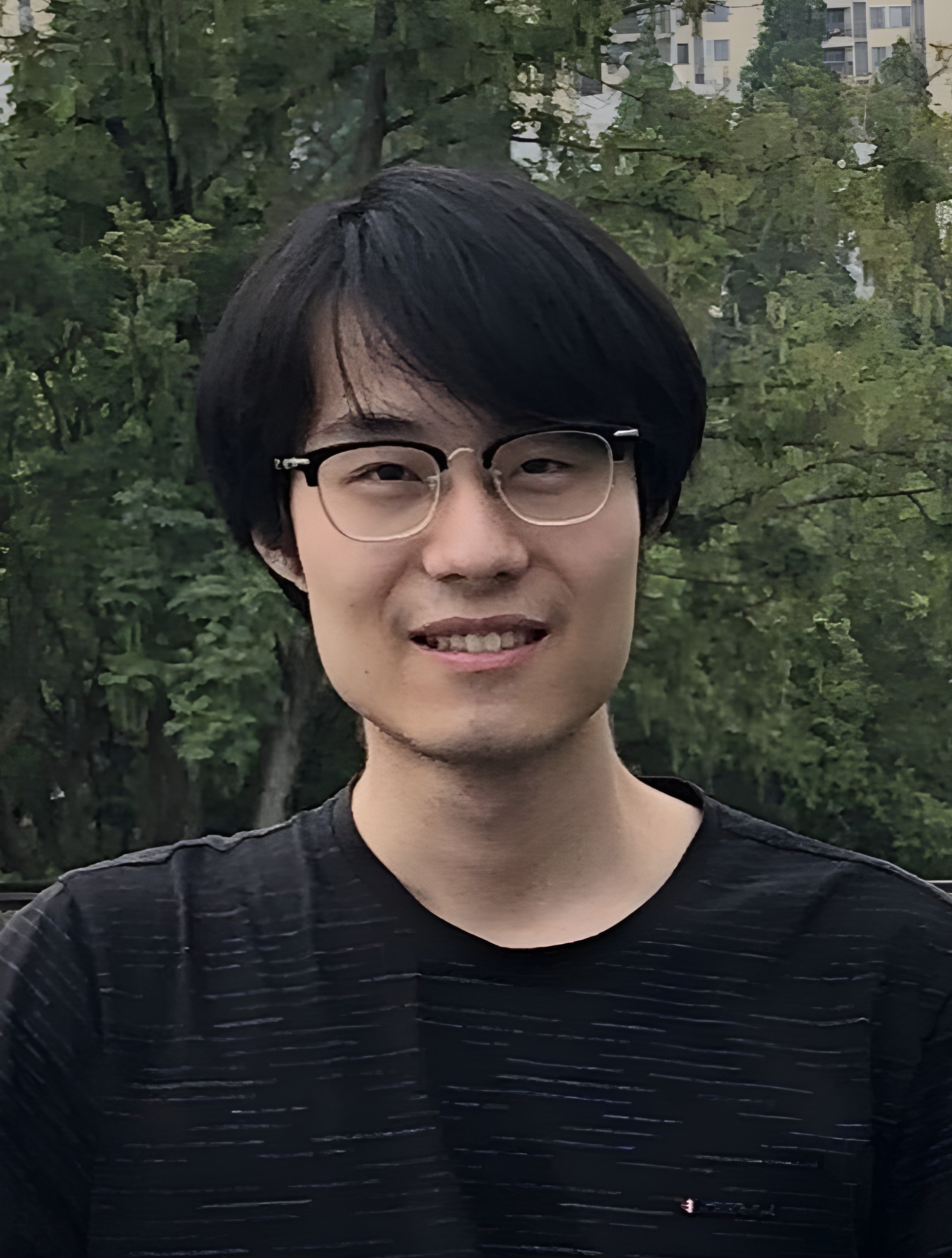}}]{Yafei Ou}
received his B.Sc. degree in Mechanical Design, Manufacturing, and Automation from the University of Electronic Science and Technology of China (UESTC), China, in 2021. He is currently pursuing a Ph.D. degree in Electrical and Computer Engineering at the University of Alberta. His research interests focus on surgical robotics control and automation.
\end{IEEEbiography}

\begin{IEEEbiography}[{\includegraphics[width=1in,height=1.25in,clip,keepaspectratio]{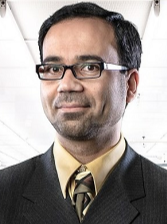}}]{Mahdi Tavakoli}
is a Professor in the Department of Electrical and Computer Engineering, University of Alberta, Canada. He received his BSc and MSc degrees in Electrical Engineering from Ferdowsi University and K.N. Toosi University, Iran, in 1996 and 1999, respectively. He received his PhD degree in Electrical and Computer Engineering from the University of Western Ontario, Canada, in 2005. In 2006, he was a post-doctoral researcher at Canadian Surgical Technologies and Advanced Robotics (CSTAR), Canada. In 2007-2008, he was an NSERC Post-Doctoral Fellow at Harvard University, USA. Dr. Tavakoli’s research interests broadly involve the areas of robotics and systems control. Specifically, his research focuses on haptics and teleoperation control, medical robotics, and image-guided surgery. Dr. Tavakoli is the lead author of Haptics for Teleoperated Surgical Robotic Systems (World Scientific, 2008). He is a Senior Member of IEEE, Specialty Chief Editor for Frontiers in Robotics and AI (Robot Design Section), and an Associate Editor for the International Journal of Robotics Research, IEEE Transactions on Medical Robotics and Bionics, IEEE Robotics and Automation Letters, IEEE TMECH/AIM Emerging Topics Focused Section, and Journal of Medical Robotics Research.
\end{IEEEbiography}

\vfill

\end{document}